\pdfoutput=1
\documentclass[10pt,twocolumn,letterpaper]{article}
\PassOptionsToPackage{table,dvipsnames}{xcolor}
\usepackage[pagenumbers]{cvpr}

\usepackage[T1]{fontenc}
\usepackage{amsthm}
\usepackage{multirow,tabularx}
\usepackage[normalem]{ulem}
\usepackage{microtype}

\makeatletter
\renewcommand{\paragraph}{\@startsection{paragraph}{4}{\z@}{.5em}{-1em}{\normalfont\normalsize\bfseries}}
\makeatother
\newtheorem{theorem}{Theorem}
\newtheorem{proposition}{Proposition}
\newtheorem{lemma}{Lemma}

\newcommand{\one}{\mathbf{1}}
\newcommand{\E}{\mathbb{E}}
\newcommand{\R}{\mathbb{R}}
\newcommand{\W}{\widehat W}
\newcommand{\td}{TMPDiff}
\newcommand{\ts}{T-Stitch}

\definecolor{apdrow}{HTML}{F3EFFB}
\newcommand{\cc}{\cellcolor{apdrow}}

\definecolor{cvprblue}{rgb}{0.21,0.49,0.74}
\usepackage[breaklinks,colorlinks,allcolors=cvprblue]{hyperref}
\hypersetup{
  pdftitle={QuoRem: One Quotient--Remainder Parent for Temporal Precision Switching in Diffusion Transformers},
  pdfauthor={Junzhao Huang, Tai Yu, Yiming Zhou, Zhenhua Xu, Yunyang Lu, Zhouzhuo Guo, Changchun Zhou}
}

\title{QuoRem: One Quotient--Remainder Parent for Temporal Precision Switching in Diffusion Transformers}
\author{Junzhao Huang$^{1}$\quad
Tai Yu$^{2}$\quad
Yiming Zhou$^{2}$\quad
Zhenhua Xu$^{3}$\\[2pt]
Yunyang Lu$^{4}$\quad
Zhouzhuo Guo$^{5}$\quad
Changchun Zhou$^{6,\dagger}$\\[6pt]
{\normalsize
$^{1}$Northeastern University\quad
$^{2}$Xi'an Jiaotong University\quad
$^{3}$Duke University}\\[-1pt]
{\normalsize
$^{4}$Columbia University\quad
$^{5}$Beijing Jiaotong University\quad
$^{6}$Peking University}\\[3pt]
{\small\texttt{huangjunzhao@mails.neu.edu.cn}\qquad $^{\dagger}$Corresponding author}}

\begin{document}
\maketitle

\begin{abstract}
Diffusion transformers (DiTs) process thousands of image tokens per step, so speedup requires quantizing both weights and activations for integer Tensor Cores.
Static W8A8 caps speedup, while static W4A4 degrades layout and detail.
Switching between W8A8 and W4A4 across steps combines 8-bit fidelity with 4-bit speed, yet existing methods hold two separate models in memory, shift 4-bit weights negative via naive bit-slicing, leave token-shared activation shifts uncentered, or use rigid one-way schedules.
We present \textbf{QuoRem}, a single-parent framework that runs W8A8 and adapter-free W4A4 from one checkpoint at $1.00$ byte per weight ($\text{B/w}$).
Offline, an offset quotient-remainder decomposition splits each 8-bit weight losslessly into a nearly unbiased 4-bit quotient plane ($\mathbf{Q}$) and a 4-bit remainder plane ($\mathbf{R}$).
Online, 4-bit steps subtract the shared token mean before quantization and restore it via one vector-matrix product across $(\mathbf{Q},\mathbf{R})$ without low-rank branches.
Along the trajectory, a one-pass Waterline profiler models early layout drift and late detail loss to unify one-switch and multi-phase schedules, while PlaneFold folds weight planes in place only at switches with negligible overhead.
Across four DiT backbones, QuoRem halves the GPU memory of dual-checkpoint baselines, reaches up to $2.18\times$ speedup over BF16 with half the steps in 4-bit, and sets the best Pareto frontier of speed and fidelity.
\end{abstract}

\section{Introduction}
\label{sec:intro}
\begin{figure*}[!t]
\centering
\includegraphics[width=\textwidth]{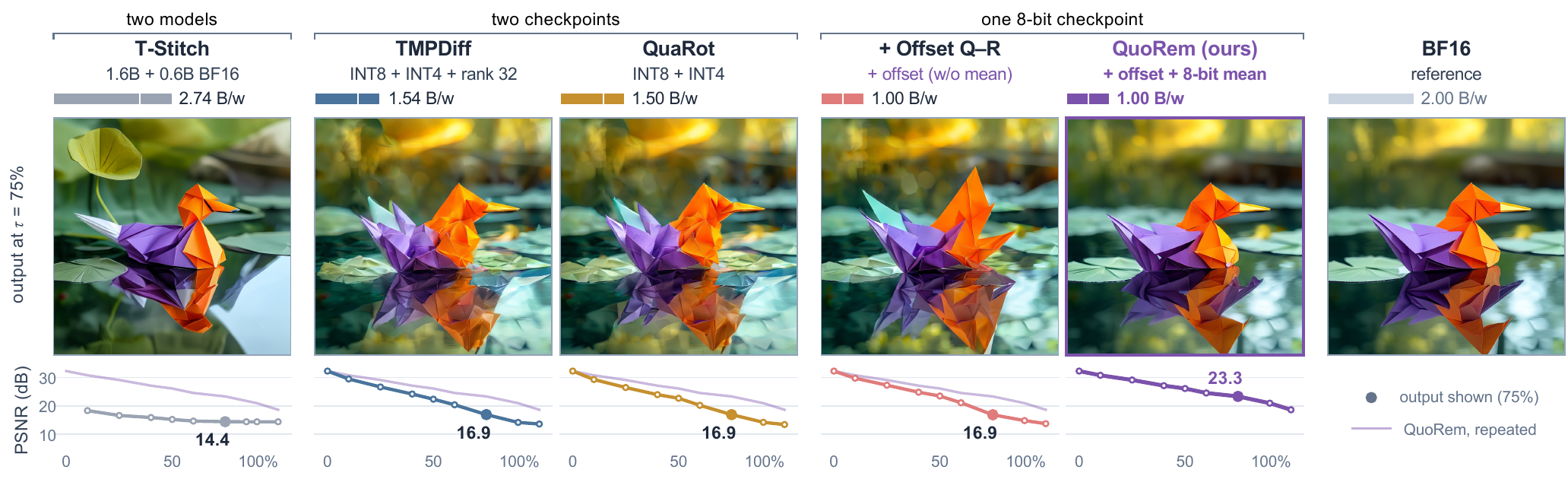}
\caption{\textbf{Single-parent precision switching on SANA-1.6B ($1024^2$).}
\textbf{Top:}~Same-seed outputs at 4-bit step share $\tau=75\%$ (filled bars: low-cost steps; $\text{B/w}$: bytes per weight). Baselines inflate memory, alter pose (\ts), or blur detail (+\,Offset Q--R), whereas \textbf{QuoRem} preserves BF16 fidelity at $1.00\text{ B/w}$.
\textbf{Bottom:}~Paired PSNR vs.\ BF16 across $\tau$ (more examples in Appendix~\ref{sec:supp_visuals}).}
\label{fig:teaser}
\end{figure*}

Diffusion transformers (DiTs) deliver state-of-the-art visual synthesis~\cite{peebles2023dit,esser2024sd3,xie2025sana,flux2024}, yet denoising thousands of image tokens in parallel makes each step compute-bound.
Single-token LLM decoding waits on memory and speeds up from weight-only W4A16~\cite{park2024anyprecisionllm}, whereas unpacking weights into BF16 matrix multiplications (GEMMs) on DiTs runs slower than BF16.
Real speedup requires quantizing both weights and activations (W8A8 or W4A4) for INT8 and INT4 Tensor Cores.
Fixed precision still forces a trade-off between speed and fidelity.
All-step W8A8 ($1.00\text{ B/w}$, bytes per weight) leaves 4-bit speedups unused, while all-step W4A4 ($0.50\text{ B/w}$) distorts global layout and fine detail (Fig.~\ref{fig:teaser}).
Because steps vary in sensitivity to quantization noise, switching between W8A8 and W4A4 along the trajectory combines 8-bit fidelity with 4-bit speed.

Prior switching methods use two separate models, which inflates memory or misaligns features.
Model stitching (T-Stitch~\cite{pan2024tstitch}) needs a smaller pretrained sibling absent in families such as FLUX.1 and passes latents across mismatched feature spaces that alter object pose.
Dual-checkpoint switching (TMPDiff~\cite{lewandowski2026tmpdiff}) holds high-precision and low-precision models in memory.
On 12B FLUX.1-dev, the pair exceeds a 24\,GB GPU and swaps over PCIe.
Slicing a 4-bit view from one 8-bit post-training quantization (PTQ) checkpoint also fails.
Dividing by $16$ with floor rounding drops positive remainders and shifts 4-bit weights negative by $7.5$ steps, while round-to-nearest-even maps $17$ codes into tie buckets that overflow a $16$-state 4-bit remainder.

Removing weight-slicing bias with a $+8$ offset (+\,Offset Q-R in Fig.~\ref{fig:teaser} and Sec.~\ref{sec:quorem}) reveals two deeper DiT bottlenecks in shared activation shifts and step placement.
Inside each step, adaptive layer normalization (AdaLN) adds a shift shared by all image tokens.
Block-Hadamard rotation mixes channels within each token to flatten outliers, yet leaves this shared mean intact ($52.5\%$ of activation energy on SANA-1.6B), wasting INT4 levels.
BF16 low-rank adapters~\cite{yao2023zeroquantv2,zhang2024lqer,li2025svdquant} leave this mean uncentered, push storage past $1.00\text{ B/w}$, and duplicate the weight residual stored in the remainder plane.
Across steps, early denoising forms global layout and amplifies injected errors up to $49\times$, while the last few large steps lack remaining damping and pass texture errors to the output (Fig.~\ref{fig:waterline_why}).
Placing all 4-bit steps early (Ascend) thus degrades layout fidelity compared with placing them late (Descend), while Descend blurs fine detail when its final steps keep residual 4-bit error.

We present \textbf{QuoRem}, one $1.00\text{ B/w}$ integer parent that runs exact-parent W8A8 steps, adapter-free W4A4 steps, and multi-phase schedules without reloading weights.
In the C++ Tensor Core engine of Nunchaku~\cite{li2025svdquant} on one RTX A5000, QuoRem halves the memory of dual-checkpoint TMPDiff, fits 12B FLUX.1-dev in $11.29\text{ GiB}$, and runs $2.18\times$ (SANA-1.6B) and $1.64\times$ (FLUX.1-dev) faster than BF16 with half the steps in 4-bit.
On SANA-1.6B, SD3.5-Medium/Large, and FLUX.1-dev, QuoRem gives the best PSNR, SSIM, and LPIPS of all compared switching methods at every 4-bit share. Our main contributions follow.
\begin{itemize}
\item \textbf{Offset Q-R and mean restoration.} Offline $+8$ offset quotient-remainder (Q-R) coding gives nearly unbiased 4-bit slicing and exact 8-bit recovery, while one online GEMV per layer restores the token mean across $(\mathbf{Q},\mathbf{R})$.
\item \textbf{Euler-Wiener Waterline scheduling.} A frequency-band model and one-pass profiler capture early drift and late blur across Ascend, Descend, and multi-phase schedules.
\item \textbf{Shared-parent runtime with PlaneFold.} In-place $(\mathbf{Q},\mathbf{R})\leftrightarrow\mathbf{Z}$ folding at switches via a single-layer swap slot lets INT8 and INT4 Tensor Cores read one bank with no per-step unpack or extra buffer.
\end{itemize}

\section{Related work}
\label{sec:related}
\paragraph{Shared and nested precision.}
Switchable networks train one model across bit-widths~\cite{jin2020adabits,yu2021anyprecision,bulat2021bitmixer}, while weight-only LLM formats pack bit-planes~\cite{park2024anyprecisionllm}, residual stacks~\cite{wang2025bitstack}, or nested integer slices~\cite{nair2025matquant,kleinegger2026matgptq} into one bank.
Weight-only quantization suits memory-bound decoding yet gives no speedup on DiTs, which accelerate only on INT8/INT4 Tensor Cores (Tab.~\ref{tab:systems}a).
Moreover, Any-Precision LLM~\cite{park2024anyprecisionllm} relies on lookup codes incompatible with integer Tensor Cores, while MatQuant~\cite{nair2025matquant} and MatGPTQ~\cite{kleinegger2026matgptq} require QAT or multi-precision GPTQ for a usable 4-bit prefix.
Naive PTQ INT8 slicing causes floor bias or round-to-nearest-even overflow (Sec.~\ref{sec:quorem}), whereas QuoRem converts a pretrained DiT without retraining into one $1.00\text{ B/w}$ parent for exact-parent W8A8 and centered W4A4 GEMMs.

\paragraph{Low-bit diffusion quantization.}
Diffusion PTQ has moved from timestep-calibrated UNets~\cite{shang2023ptq4dm,li2023qdiffusion,he2023ptqd,he2024efficientdm} to DiT smoothing and temporal calibration~\cite{wu2024ptq4dit,zhao2024mixdq,zhao2025viditq,chen2025qdit,zhang2026segquant}.
W4A4 PTQ uses smoothing or rotation~\cite{xiao2023smoothquant,ashkboos2024quarot,liu2025spinquant}, channel shifting~\cite{wei2023outlier}, centering~\cite{federici2025hadanorm,jovanovic2026centriq}, or BF16 low-rank branches~\cite{yao2023zeroquantv2,zhang2024lqer,li2025svdquant}.
These PTQ methods fix precision across steps, and BF16 branches exceed $1.00\text{ B/w}$ while duplicating $\mathbf{R}$ (Sec.~\ref{sec:deletion}); TASQ~\cite{han2026tasq} varies stage-wise weight bits from one buffer, yet trains LSB masks for bit-serial hardware rather than PTQ W8A8/W4A4 Tensor Cores.
Unlike fixed-precision HadaNorm~\cite{federici2025hadanorm} and per-token CentriQ~\cite{jovanovic2026centriq}, QuoRem centers across tokens and restores the shared mean across $(\mathbf{Q},\mathbf{R})$ at $r=0$ (Sec.~\ref{sec:mean}).

\paragraph{Temporal switching and trajectory acceleration.}
Fast solvers~\cite{lu2022dpmsolver}, feature caching~\cite{ma2024deepcache}, and token pruning~\cite{sun2026jit} cut steps or skip blocks and are orthogonal to precision switching at a fixed step count $N$.
At fixed $N$, prior methods adapt step scales~\cite{so2023tdq,huang2024tfmqdm}, route across expert or slimmable denoisers~\cite{balaji2022ediffi,liu2023omsdpm,yang2024ddsm}, stitch siblings early (Ascend)~\cite{pan2024tstitch}, or greedily rank step upcasting across dual checkpoints in TMPDiff~\cite{lewandowski2026tmpdiff}, which resembles Descend.
Dual models hold $1.50$ to $2.74\text{ B/w}$ in memory, and TMPDiff needs $29.6\text{ GiB}$ on 12B FLUX.1-dev, exceeding a 24\,GB GPU.
One-way ranking also misses late detail loss from residual 4-bit error (Sec.~\ref{sec:scheduling}), whereas QuoRem runs Ascend, Descend, and Waterline from one $1.00\text{ B/w}$ parent.

\section{Methodology}
\label{sec:method}
Consider an $N$-step denoising trajectory $z_j=z_{j-1}-\Delta t_j v^{(\pi_j)}(z_{j-1},t_{j-1},y)$ with step size $\Delta t_j$ and denoiser velocity $v^{(\pi_j)}$, where schedule $\pi\in\{\text{W8A8},\text{W4A4}\}^N$ runs $k$ steps ($\tau=k/N$) in W4A4 and the remaining $N-k$ steps in W8A8.
Our goal is to keep the final image $z_N$ close to BF16 within a target latency budget while storing only $1.00\text{ B/w}$ of weights on the GPU.
QuoRem achieves this by combining an offset quotient-remainder weight parent (Sec.~\ref{sec:quorem}), rank-zero activation mean restoration (Sec.~\ref{sec:mean}), sensitivity-guided trajectory scheduling (Sec.~\ref{sec:scheduling}), and in-place weight folding at switches (Sec.~\ref{sec:runtime}).

\subsection{QuoRem: Offset quotient-remainder parent}
\label{sec:quorem}
Storing separate W8 and W4 checkpoints costs $1.50\text{ B/w}$, yet slicing a 4-bit weight directly from a signed INT8 code $z\in[-128,127]$ (step scale $s>0$) fails in post-training quantization.
Dividing $z$ by $16$ with floor rounding ($q_{\text{floor}}=\lfloor z/16\rfloor$) drops non-negative remainders $\{0,\ldots,15\}$ (average $+7.5$) and pulls 4-bit weights down by $7.5s$.
Round-to-nearest-even ($q_{\text{rne}}=\operatorname{RNE}(z/16)$) sends both ties $16q-8$ and $16q+8$ to the same even quotient $q$, putting $17$ inputs into one bucket and overflowing a $16$-state 4-bit remainder.

\begin{figure}[!t]
\centering
\includegraphics[width=\columnwidth]{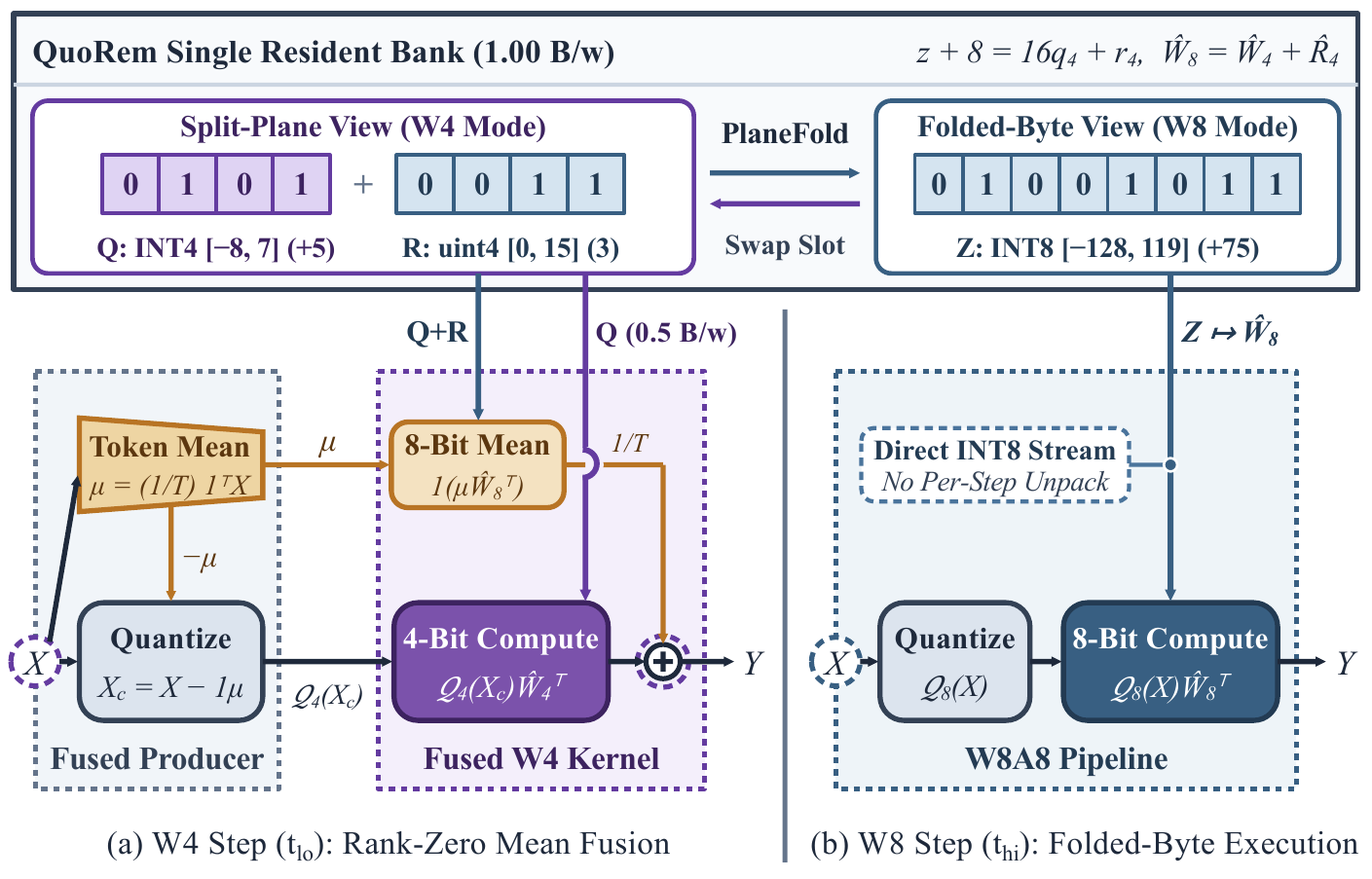}
\caption{\textbf{QuoRem architecture and PlaneFold layout.}
\textbf{Top:}~One $1.00\text{ B/w}$ bank maps split planes $[\mathbf{Q};\mathbf{R}]$ bijectively to INT8 bytes $\mathbf{Z}$ via in-place PlaneFold at precision switches.
\textbf{Bottom-left:}~W4 streams $\mathbf{Q}$ on INT4 Tensor Cores and restores the token mean $\mu$ across $\mathbf{Q}+\mathbf{R}$ ($r=0$).
\textbf{Bottom-right:}~W8 directly streams $\mathbf{Z}$ without per-step unpacking.}
\label{fig:quorem_arch}
\end{figure}

QuoRem fixes both with an offset quotient-remainder coding (Offset Q-R) that clips the INT8 parent $\mathbf{Z}$ to $z\in\{-128,\ldots,119\}$ ($248$ levels) and adds $+8$ before floor division (Fig.~\ref{fig:quorem_arch}, top).
Mapping $z+8\in[-120,127]$ splits every code without overflow into a signed INT4 quotient $\mathbf{Q}$ ($q_4\in[-8,7]$) and an unsigned 4-bit remainder $\mathbf{R}$ ($r_4\in[0,15]$) as
\begin{equation}
q_4=\left\lfloor\frac{z+8}{16}\right\rfloor,\quad
r_4=z+8-16q_4,\quad
\W_8=\W_4+\widehat R_4,
\label{eq:quorem}
\end{equation}
where $\W_8=s\mathbf{Z}$, $\W_4=16s\mathbf{Q}$, and $\widehat R_4=s(\mathbf{R}-8)$.
Because $\W_8-\W_4=s(r_4-8)$ centers the dropped remainder on $r_4-8\in\{-8,\ldots,7\}$, the $+8$ offset shrinks average slicing bias $15\times$ from $7.5s$ to $0.5s$.
Using all $16$ levels $[-8,7]$ at step $16s$ also gives finer resolution than symmetric $[-7,7]$ INT4, beating standalone INT4 fidelity while $(\mathbf{Q},\mathbf{R})$ recovers $\W_8$ exactly at $1.00\text{ B/w}$ (Tab.~\ref{tab:parent_ablation}; Appendix~\ref{sec:supp_quorem}).

\subsection{Rank-zero cross-plane mean restoration}
\label{sec:mean}\label{sec:policy}
In DiTs, AdaLN conditioning adds the same channel shift across all $T$ image tokens.
For rotated activations $X\in\R^{T\times D}$ under an orthogonal block-Hadamard transform $\widetilde H_B=I_{D/B}\otimes H_B$, rotating channels inside each token does not change the cross-token average $\mu=T^{-1}\one^\top X\in\R^{1\times D}$ ($\one\in\R^{T\times 1}$).
Quantizing $X$ directly wastes INT4 dynamic range on $\mu$ instead of spatial variations, while BF16 low-rank adapters leave $\mu$ uncentered, break the $1.00\text{ B/w}$ budget, and duplicate the residual stored in $\widehat R_4$.

We cancel $\mu$ with zero low-rank adapters ($r=0$) by subtracting it to form $X_c=X-\one\mu$ before per-token INT4 quantization $\mathcal Q_4(\cdot)$ and adding back $\one(\mu\W_8^\top)$ across $(\mathbf{Q},\mathbf{R})$ via $\W_8=\W_4+\widehat R_4$ (8-Bit Mean, Fig.~\ref{fig:quorem_arch}, bottom-left) as
\begin{equation}
Y_{\text{online}}=\mathcal Q_4(X_c)\W_4^\top+\one(\mu\W_8^\top)+b.
\label{eq:m8}
\end{equation}
With centered rounding error $E_4=\mathcal Q_4(X_c)-X_c$, substituting $X=X_c+\one\mu$ cancels the shared-mean remainder error $\one\mu\widehat R_4^\top$ against the 8-bit target $X\W_8^\top+b$, giving
\begin{equation}
Y_{\text{online}}-(X\W_8^\top+b)=E_4\W_4^\top-X_c\widehat R_4^\top.
\label{eq:meanerror}
\end{equation}
Computing $\mu\W_8^\top\in\R^{1\times D_{\text{out}}}$ across $(\mathbf{Q},\mathbf{R})$ takes just one vector-matrix product (GEMV, $O(1/T)$ of the GEMM cost) and leaves zero-mean residuals ($\one^\top X_c\widehat R_4^\top=0$).
Using $\W_4$ alone ($\mu\W_4^\top$, 4-Bit Mean) misses $\mathbf{R}$ and leaves a constant bias $-\one(\mu\widehat R_4^\top)$ on all $T$ tokens (Tab.~\ref{tab:mean_ablation}).
For skewed post-GELU FF-out layers with predictable step means passing validation gates (Sec.~\ref{sec:setup}), Static FF-Shift folds a calibrated step mean $\bar\mu_t$ offline into $b_t'=b+\bar\mu_t\W_8^\top$ and runs $Y_{\text{static},t}=\mathcal Q_4(X_t-\one\bar\mu_t)\W_4^\top+b_t'$ without online reduction (Appendix~\ref{sec:supp_mean}).

\subsection{Euler-Wiener Waterline scheduling}
\label{sec:scheduling}
\begin{figure}[!t]
\centering
\includegraphics[width=\columnwidth]{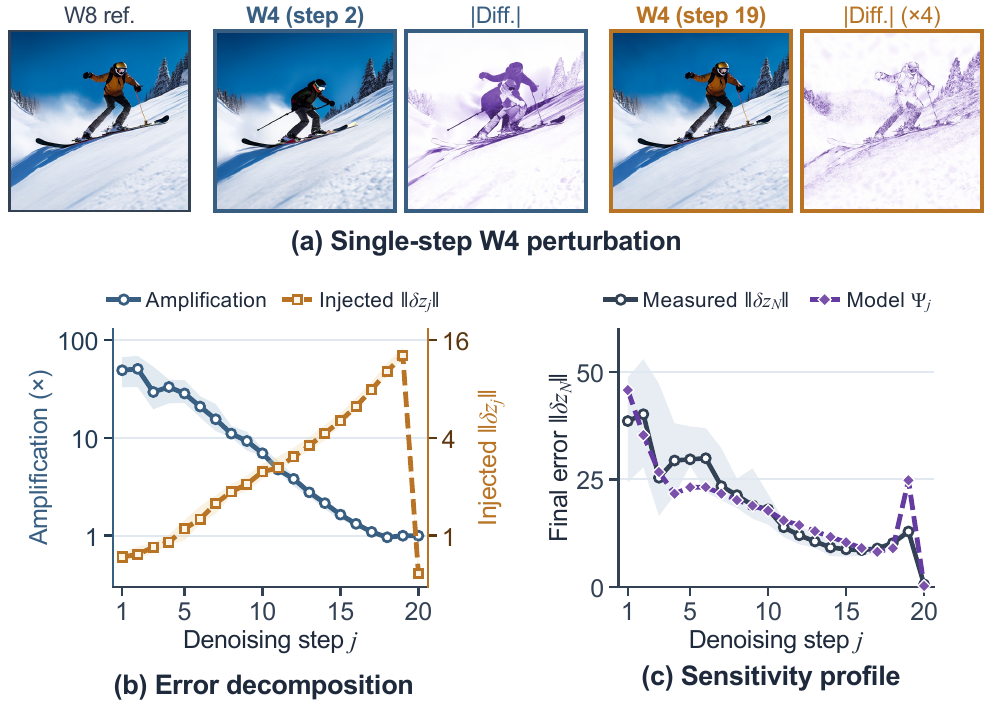}
\caption{\textbf{Two-ended step sensitivity on SANA-1.6B} (one W4 step per W8 run, 16 prompts).
\textbf{(a)}~Outputs and $|\text{Diff.}|$ vs.\ W8: early W4 shifts layout; late W4 blurs texture.
\textbf{(b)}~Amplification vs.\ injected $\|\delta z_j\|$ (median, IQR): $49\times$ early gain and late uncontracted injection.
\textbf{(c)}~One-pass $\Psi_j$ tracks $\|\delta z_N\|$ ($\rho=0.91$).}
\label{fig:waterline_why}
\end{figure}

Mean restoration removes the uniform ($f=0$) offset per step, yet spatial 4-bit errors ($f\ge 1$) still propagate along $z_t=(1-t)x_0+t\varepsilon$ ($t_0=1>\cdots>t_N=0$).
Switching one step to W4 in a W8 run shows why placing all $k$ W4 steps early (Ascend) or late (Descend) can fail, as early W4 shifts layout while late W4 blurs detail (Fig.~\ref{fig:waterline_why}a).

To explain both effects, we split final error into the step-$j$ injection $\delta z_j=-\Delta t_j\delta v_j$ ($\Delta t_j=t_{j-1}-t_j$) and the survival gain $\kappa_{j,f}^{\text{E}}$ scaling it to $t_N=0$ (Fig.~\ref{fig:waterline_why}b).
Modeling clean frequency band $f$ as Gaussian with variance $\sigma_f^2$ (large for coarse layout, small for fine texture) under an MMSE Wiener denoiser, Euler compounding bounds this gain by
\begin{equation}
\kappa_{j,f}^{\text{E}}\le\sqrt{\frac{\sigma_f^2}{(1-t_j)^2\sigma_f^2+t_j^2}}.
\label{eq:ew}
\end{equation}
Equation~\eqref{eq:ew} explains both arms in Fig.~\ref{fig:waterline_why}b,c.
(i)~\emph{Early layout drift.} Low-frequency layout bands with large $\sigma_f^2$ amplify early errors by $\kappa_{j,f}^{\text{E}}\approx(1-t_j)^{-1}>1$ under high noise variance $t_{j-1}^2$.
(ii)~\emph{Middle damping and late exposure.} High-frequency texture bands with small $\sigma_f^2$ contract mid-trajectory errors ($\kappa_{j,f}^{\text{E}}\ll 1$) into a low-error valley, yet almost no damping remains as $t_j\to 0$ ($\kappa_{j,f}^{\text{E}}\to 1$) where large relative step sizes ($\Delta t_j/t_{j-1}$) expose 4-bit texture blur.

To avoid $O(N)$ full rollouts, one fast shadow pass evaluates the resident W4 view along the W8 trajectory to measure step error $\mathcal{E}_j^{\text{rel}}=\E\|\delta v_j\|_2^2/\E\|v_{8,j}\|_2^2$ ($\delta v_j=v_4-v_8$).
Matching both regimes, we combine early layout term $a_j$ and late detail term $d_j$ ($\lambda,\eta_0,\varepsilon_0>0$) into step score
\begin{equation}
\Psi_j(\lambda)=\underbrace{\frac{\Delta t_j}{1-t_j+\eta_0}t_{j-1}^2\mathcal{E}_j^{\text{rel}}}_{a_j} + \lambda\underbrace{\left(\frac{\Delta t_j}{t_{j-1}+\varepsilon_0}\right)^2(\mathcal{E}_j^{\text{rel}})^2}_{d_j}.
\label{eq:direction}
\end{equation}
As Fig.~\ref{fig:waterline_why}c shows, $\Psi_j(\lambda)$ closely tracks true terminal distortion.
For $k$ W4 steps, Waterline sorts $\{\Psi_j(\lambda)\}_{j=1}^N$ in ascending order ($j_1,\dots,j_N$, ties broken by $j$) and picks the $k$ lowest-score steps below threshold $\Psi_{(k)}=\Psi_{j_k}(\lambda)$, forming nested sets $S^\star(1)\subset\cdots\subset S^\star(N)$ given by
\begin{equation}
S^\star(k)=\{j_1,\dots,j_k\},\quad \Psi_{j_i}(\lambda)\le\Psi_{(k)}.
\label{eq:waterline}
\end{equation}
Raising $\Psi_{(k)}$ assigns low-sensitivity valley steps to W4 while keeping early layout peaks and tail spikes in W8.
To avoid fitting calibration noise on small prompt sets, Waterline falls back to single-switch Descend unless $S^\star(k)$ reduces the normalized score sum over Descend by $\Delta_{\text{WL}}\ge\delta_{\min}$ (normalized by $\sum_{i=1}^N\Psi_i(\lambda)$, Sec.~\ref{sec:setup}; Appendix~\ref{sec:supp_waterline}).

\begin{figure*}[!t]
\centering
\includegraphics[width=\textwidth]{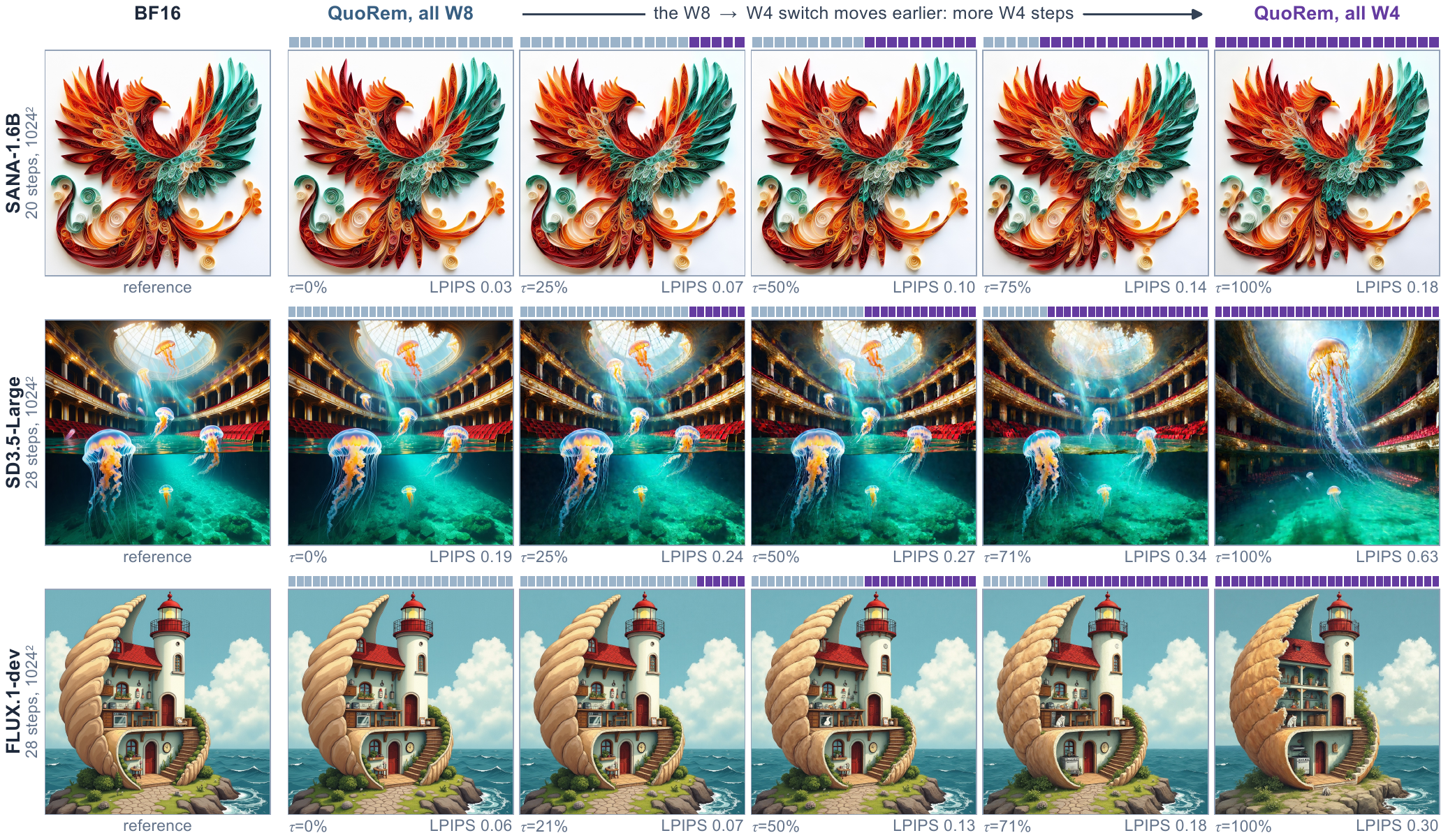}
\caption{\textbf{Progression across W8$\to$W4 step shares $\tau$} (\textbf{Descend}, $1024^2$, shared seed; labels: per-image LPIPS vs.\ BF16).
At $1.00\text{ B/w}$, QuoRem keeps layout and detail up to $\tau\approx 75\%$; Tab.~\ref{tab:systems} gives the SANA and FLUX speed at each $\tau$ (more examples in Appendix~\ref{sec:supp_visuals}).}
\label{fig:switch_gallery}
\end{figure*}

\subsection{In-place PlaneFold and Tensor Core runtime}
\label{sec:runtime}
INT4 Tensor Cores need contiguous quotients $\mathbf{Q}$ in split layout $(\mathbf{Q},\mathbf{R})$, whereas INT8 Tensor Cores need combined bytes $\mathbf{Z}$.
Unpacking $\mathbf{Z}=16\mathbf{Q}+\mathbf{R}-8$ on every W8 step wastes compute, while storing a second INT8 copy doubles memory to $2.00\text{ B/w}$.
Because schedules switch only a few times per run and $(q_4,r_4)\leftrightarrow z=16q_4+r_4-8$ is a size-preserving bijection, In-Place Plane Folding (PlaneFold) rearranges each $1.00\text{ B/w}$ buffer in place only when precision switches, using a scratch buffer sized for a single layer (Fig.~\ref{fig:quorem_arch}, top).
(i)~\emph{Split-plane W4 view (Fig.~\ref{fig:quorem_arch}, bottom-left).} At step time $t_{\mathrm{lo}}$, fused kernels rotate, center, and quantize $X_c$ while INT4 Tensor Cores stream $\mathbf{Q}$ alongside $\one(\mu\W_8^\top)$.
(ii)~\emph{Folded-byte W8 view (Fig.~\ref{fig:quorem_arch}, bottom-right).} At switches, a fast SIMD pass folds $(\mathbf{Q},\mathbf{R})\leftrightarrow\mathbf{Z}$ through the single-layer scratch slot so W8 steps (time $t_{\mathrm{hi}}$) stream $\mathbf{Z}$ directly into INT8 Tensor Cores with zero per-step unpacking.
Across $N-k$ W8 steps and $k$ W4 steps, trajectory latency (Appendix~\ref{sec:supp_planefold}) follows
\begin{equation}
\ell(\pi)=(N-k)t_{\mathrm{hi}}+kt_{\mathrm{lo}}+H(\pi),
\label{eq:latency}
\end{equation}
where $H(\pi)=n_{8\to 4}h_{8\to 4}+n_{4\to 8}h_{4\to 8}+H_0$ collects the small PlaneFold overhead across $n_{8\to 4}$ and $n_{4\to 8}$ switches.

\begin{figure}[!t]
\centering
\includegraphics[width=0.98\columnwidth]{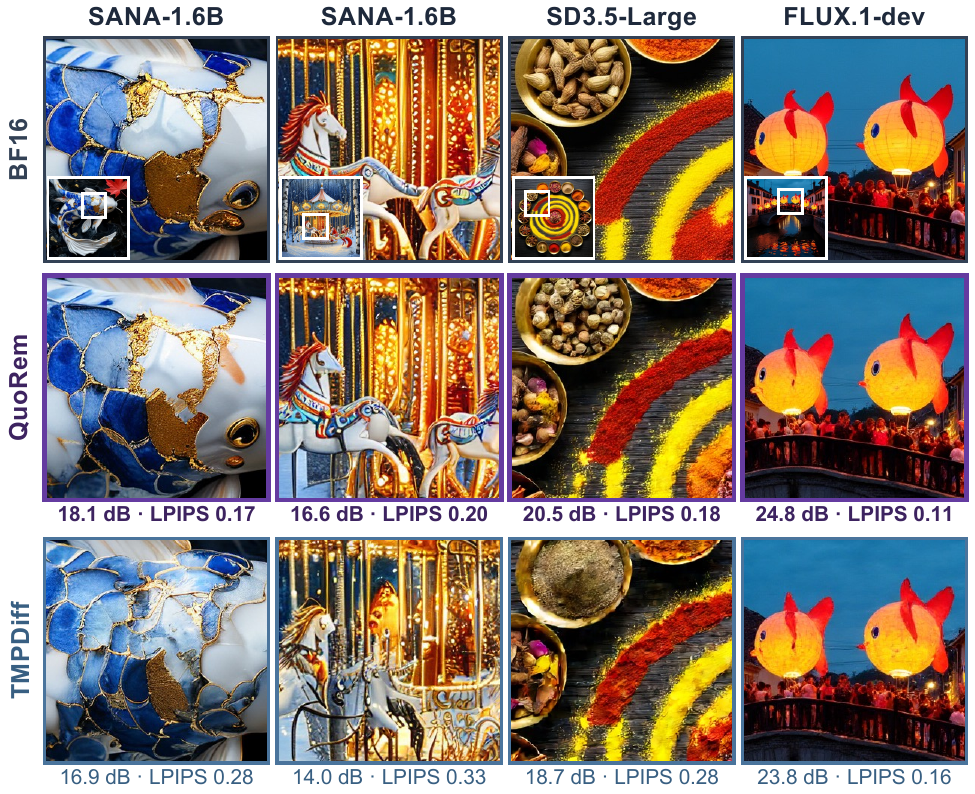}
\caption{\textbf{Pixel-level crops at $\tau=50\%$} (\textbf{Descend}, native resolution; bottom: full-image PSNR/LPIPS vs.\ BF16).
QuoRem preserves gold seams and carousel details on SANA-1.6B, spices on SD3.5-Large, and lantern goldfish on FLUX.1-dev blurred by \td\ (extended galleries in Appendix~\ref{sec:supp_visuals}).}
\label{fig:detail}
\end{figure}

\begin{table*}[!t]
\centering
\caption{\textbf{Main switching comparison across W4 step shares $\tau$} (controlled shared-parent track, 998 MS-COCO prompts, RTX A5000; \textbf{bold}: best, \underline{underlined}: second).}
\label{tab:tmpdiff}\label{tab:tstitch}
\footnotesize\setlength{\tabcolsep}{8.5pt}\renewcommand{\arraystretch}{0.84}\setlength{\aboverulesep}{0.18ex}\setlength{\belowrulesep}{0.12ex}\setlength{\ULdepth}{0.95pt}\renewcommand{\ULthickness}{0.36pt}\let\underline\uline
\newcommand{\rtop}{\rule{0pt}{2.02ex}}
\newcommand{\bcell}[2]{\begin{tabular}{@{}c@{}}\textbf{#1}\\[-0.25ex]{\scriptsize(#2)}\end{tabular}}
\begin{tabularx}{\textwidth}{@{}ccXrrrrrrr@{}}
\toprule
\multirow{2}{*}[-0.35ex]{\textbf{Backbone}} & \multirow{2}{*}[-0.35ex]{\textbf{$\tau$}} & \multirow{2}{*}[-0.35ex]{\textbf{Method}} & \multirow{2}{*}[-0.35ex]{\textbf{B/w} $\downarrow$} & \multicolumn{4}{c}{\textbf{Trajectory Fidelity (vs.\ BF16)}} & \multicolumn{2}{c@{}}{\textbf{Generation Quality}}\\
\cmidrule(lr){5-8}\cmidrule(l){9-10}
\rtop & & & & \textbf{PSNR} $\uparrow$ & \textbf{SSIM} $\uparrow$ & \textbf{LPIPS} $\downarrow$ & \textbf{FID}$_{\text{BF16}}$ $\downarrow$ & \textbf{FID} $\downarrow$ & \textbf{IR} $\uparrow$\\
\midrule
\multirow{10}{*}[-2.6ex]{\bcell{SANA-1.6B}{$1024^2$, 20 steps}}
 & \multirow{2}{*}[-0.35ex]{0\%}
   & \rtop BF16 & 2.00 & -- & -- & -- & -- & 71.74 & 1.007\\
 & & \textbf{QuoRem}\cc & \cc\textbf{1.00} & \cc\textbf{26.31} & \cc\textbf{0.907} & \cc\textbf{0.052} & \cc\textbf{9.34} & \cc\textbf{71.24} & \cc\textbf{1.014}\\
\cmidrule(lr){2-10}
 & \multirow{4}{*}[-0.7ex]{50\%}
   & \rtop \ts & 2.74 & 12.45 & 0.465 & 0.513 & 39.96 & 72.65 & \textbf{1.027}\\
 & & \td & 1.54 & \underline{19.45} & \underline{0.674} & \underline{0.199} & \underline{29.84} & 71.20 & 0.975\\
 & & QuaRot & \underline{1.50} & 19.43 & 0.673 & \underline{0.199} & 30.08 & \textbf{70.61} & 0.965\\
 & & \textbf{QuoRem}\cc & \cc\textbf{1.00} & \cc\textbf{21.96} & \cc\textbf{0.784} & \cc\textbf{0.107} & \cc\textbf{16.75} & \cc\underline{71.06} & \cc\underline{1.015}\\
\cmidrule(lr){2-10}
 & \multirow{4}{*}[-0.7ex]{100\%}
   & \rtop \ts & 2.74 & 12.22 & \underline{0.448} & 0.533 & \underline{42.38} & \underline{72.23} & \underline{1.004}\\
 & & \td & 1.54 & \underline{12.93} & 0.426 & \underline{0.530} & 66.79 & 80.19 & 0.616\\
 & & QuaRot & \underline{1.50} & \underline{12.93} & 0.425 & 0.531 & 67.61 & 80.41 & 0.606\\
 & & \textbf{QuoRem}\cc & \cc\textbf{1.00} & \cc\textbf{17.24} & \cc\textbf{0.646} & \cc\textbf{0.253} & \cc\textbf{28.84} & \cc\textbf{71.22} & \cc\textbf{1.017}\\
\midrule
\multirow{8}{*}[-2.1ex]{\bcell{SD3.5-Medium}{$512^2$, 28 steps}}
 & \multirow{2}{*}[-0.35ex]{0\%}
   & \rtop BF16 & 2.00 & -- & -- & -- & -- & 70.91 & 1.028\\
 & & \textbf{QuoRem}\cc & \cc\textbf{1.00} & \cc\textbf{22.91} & \cc\textbf{0.868} & \cc\textbf{0.110} & \cc\textbf{16.36} & \cc\textbf{71.00} & \cc\textbf{1.024}\\
\cmidrule(lr){2-10}
 & \multirow{3}{*}[-0.5ex]{50\%}
   & \rtop \td & 1.52 & \underline{19.32} & \underline{0.697} & \underline{0.269} & \underline{48.95} & \underline{74.70} & \underline{0.935}\\
 & & QuaRot & \underline{1.50} & 18.36 & 0.625 & 0.371 & 78.89 & 86.32 & 0.736\\
 & & \textbf{QuoRem}\cc & \cc\textbf{1.00} & \cc\textbf{19.71} & \cc\textbf{0.764} & \cc\textbf{0.202} & \cc\textbf{30.56} & \cc\textbf{70.24} & \cc\textbf{0.995}\\
\cmidrule(lr){2-10}
 & \multirow{3}{*}[-0.5ex]{100\%}
   & \rtop \td & 1.52 & \underline{13.45} & \underline{0.518} & \underline{0.501} & \underline{65.41} & \underline{77.29} & \underline{0.800}\\
 & & QuaRot & \underline{1.50} & 12.66 & 0.438 & 0.602 & 116.45 & 106.32 & 0.193\\
 & & \textbf{QuoRem}\cc & \cc\textbf{1.00} & \cc\textbf{14.06} & \cc\textbf{0.551} & \cc\textbf{0.451} & \cc\textbf{59.31} & \cc\textbf{74.46} & \cc\textbf{0.863}\\
\midrule
\multirow{10}{*}[-2.6ex]{\bcell{SD3.5-Large}{$1024^2$, 28 steps}}
 & \multirow{2}{*}[-0.35ex]{0\%}
   & \rtop BF16 & 2.00 & -- & -- & -- & -- & 68.79 & 1.047\\
 & & \textbf{QuoRem}\cc & \cc\textbf{1.00} & \cc\textbf{19.76} & \cc\textbf{0.820} & \cc\textbf{0.191} & \cc\textbf{22.95} & \cc\textbf{68.40} & \cc\textbf{1.056}\\
\cmidrule(lr){2-10}
 & \multirow{4}{*}[-0.7ex]{50\%}
   & \rtop \ts & 2.61 & 10.67 & 0.488 & 0.618 & \underline{46.23} & \underline{69.35} & \textbf{1.089}\\
 & & \td & 1.52 & 18.51 & \underline{0.697} & \underline{0.339} & 53.36 & 76.00 & 1.006\\
 & & QuaRot & \underline{1.50} & \underline{18.52} & \underline{0.697} & 0.343 & 55.46 & 77.29 & 0.998\\
 & & \textbf{QuoRem}\cc & \cc\textbf{1.00} & \cc\textbf{19.14} & \cc\textbf{0.789} & \cc\textbf{0.230} & \cc\textbf{29.08} & \cc\textbf{68.00} & \cc\underline{1.048}\\
\cmidrule(lr){2-10}
 & \multirow{4}{*}[-0.7ex]{100\%}
   & \rtop \ts & 2.61 & 10.62 & 0.479 & 0.635 & \textbf{47.98} & \textbf{69.52} & \textbf{1.052}\\
 & & \td & 1.52 & \underline{12.51} & \underline{0.519} & \underline{0.587} & 76.00 & 85.60 & 0.934\\
 & & QuaRot & \underline{1.50} & 12.34 & 0.514 & 0.597 & 79.11 & 87.79 & 0.897\\
 & & \textbf{QuoRem}\cc & \cc\textbf{1.00} & \cc\textbf{13.20} & \cc\textbf{0.579} & \cc\textbf{0.495} & \cc\underline{51.72} & \cc\underline{71.37} & \cc\underline{1.023}\\
\midrule
\multirow{8}{*}[-2.1ex]{\bcell{FLUX.1-dev}{$1024^2$, 28 steps}}
 & \multirow{2}{*}[-0.35ex]{0\%}
   & \rtop BF16 & 2.00 & -- & -- & -- & -- & 72.07 & 1.049\\
 & & \textbf{QuoRem}\cc & \cc\textbf{1.00} & \cc\textbf{27.76} & \cc\textbf{0.918} & \cc\textbf{0.080} & \cc\textbf{10.20} & \cc\textbf{72.01} & \cc\textbf{1.051}\\
\cmidrule(lr){2-10}
 & \multirow{3}{*}[-0.5ex]{50\%}
   & \rtop \td & 1.53 & \underline{25.71} & \underline{0.875} & \underline{0.127} & \underline{17.61} & \underline{71.38} & \underline{1.045}\\
 & & QuaRot & \underline{1.50} & 25.55 & 0.874 & 0.129 & 17.90 & \textbf{71.36} & 1.043\\
 & & \textbf{QuoRem}\cc & \cc\textbf{1.00} & \cc\textbf{26.42} & \cc\textbf{0.894} & \cc\textbf{0.101} & \cc\textbf{13.53} & \cc 71.58 & \cc\textbf{1.048}\\
\cmidrule(lr){2-10}
 & \multirow{3}{*}[-0.5ex]{100\%}
   & \rtop \td & 1.53 & \underline{17.36} & \underline{0.699} & 0.379 & 36.44 & \textbf{70.68} & \underline{1.012}\\
 & & QuaRot & \underline{1.50} & \underline{17.36} & \underline{0.699} & \underline{0.378} & \underline{36.37} & \underline{71.13} & 1.011\\
 & & \textbf{QuoRem}\cc & \cc\textbf{1.00} & \cc\textbf{18.63} & \cc\textbf{0.744} & \cc\textbf{0.310} & \cc\textbf{30.23} & \cc 72.04 & \cc\textbf{1.027}\\
\bottomrule
\end{tabularx}
\end{table*}

\section{Experiments}
\label{sec:experiments}

\subsection{Experimental setup}
\label{sec:setup}
\paragraph{Models and metrics.}
We evaluate four DiTs, namely $1024^2$ SANA-1.6B~\cite{xie2025sana} ($N=20$, CFG 4.5), $512^2$ SD3.5-Medium and $1024^2$ SD3.5-Large~\cite{esser2024sd3} ($N=28$, CFG 7.0), and $1024^2$ FLUX.1-dev~\cite{flux2024} (12B, $N=28$, guidance 3.5).
Following~\cite{li2025svdquant,lewandowski2026tmpdiff}, we test on $998$ MS-COCO captions~\cite{lin2014coco} ($256$ in ablations).
Paired fidelity measures pixel and perceptual similarity to the BF16 image from the same prompt and random seed via PSNR, SSIM, LPIPS~\cite{zhang2018lpips}, and $\text{FID}_{\text{BF16}}$ (FID against matched BF16), while standalone quality uses FID~\cite{heusel2017fid} and ImageReward~(IR)~\cite{xu2023imagereward}.

\paragraph{Two-track baseline protocol.}
We separate algorithmic controls from system benchmarks, measuring all latency, speedup, and peak memory on one exclusive NVIDIA RTX A5000 (24\,GB).
(i)~\emph{Controlled switching (Tabs.~\ref{tab:tmpdiff}, \ref{tab:schedule_comparison}).} In our shared-parent executor ($B=64$ on SANA, $128$ on SD3.5/FLUX, SmoothQuant $\alpha=0.8$~\cite{xiao2023smoothquant}), we compare QuoRem ($r=0$, $1.00\text{ B/w}$) at matched $\tau$ against sibling stitching (\ts~\cite{pan2024tstitch}), dual-checkpoint W8/W4 with rank-32 branches (\td~\cite{lewandowski2026tmpdiff,li2025svdquant,yao2023zeroquantv2,zhang2024lqer}), and dual-checkpoint symmetric INT4 (QuaRot~\cite{ashkboos2024quarot}).
(ii)~\emph{System efficiency (Tabs.~\ref{tab:systems}, \ref{tab:handoff}, Fig.~\ref{fig:trajectory}).} On SANA-1.6B and FLUX.1-dev in C++ Nunchaku~\cite{li2025svdquant}, we deploy QuoRem ($r=0$, $B=64$, clip $0.9$) with PlaneFold against $\text{batch}=1$ SVDQuant (INT4 $r=32$), NF4 (W4A16), TorchAO INT8, LLM.int8(), same-engine unrotated per-row Nunchaku W8A8, T-Stitch ($\text{0.6B}\to\text{1.6B}$), and TMPDiff (BF16+SVDQuant~\cite{lewandowski2026tmpdiff}).

\paragraph{Implementation details.}
QuoRem stores a $1.00\text{ B/w}$ $(\mathbf{Q},\mathbf{R})$ payload ($+0.031$/$+0.016\text{ B/w}$ BF16 scales at $B=64/128$, INT8 clip $119$, and $304$ $(\mathbf{Q},\mathbf{R})$ plus $76$ W4A16 AdaLN layers on FLUX C++).
On skewed post-GELU FF-out layers, Static FF-Shift is enabled only when its velocity error $\text{MSE}_v$ on held-out BF16 states stays within $1.10\times$ online 8-Bit Mean and 16-prompt PSNR/LPIPS do not drop, selecting Static FF-Shift on SD3.5-M/L and online 8-Bit Mean on SANA/FLUX.
At $\tau=50\%$, W8/W4 baselines use Descend (beating TMPDiff's greedy schedule on all four DiTs), while QuoRem runs Waterline profiled on $4$ prompts ($8$ runs, $\lambda=0.35,\varepsilon_0=0.08,\eta_0=10^{-6},\delta_{\min}=0.05$).

\subsection{Main results}
\label{sec:schedules}
\textbf{QuoRem} attains the best PSNR, SSIM, and LPIPS in every backbone and $\tau$ cell of Tab.~\ref{tab:tmpdiff} and the lowest $\text{FID}_{\text{BF16}}$ in seven of the eight switched cells, all at $1.00\text{ B/w}$.
Without a second checkpoint, QuoRem lowers $\tau=50\%$ $\text{FID}_{\text{BF16}}$ on SD3.5-Medium from $48.95$ to $30.56$ and raises all-W4 PSNR on SANA-1.6B by $4.31\text{ dB}$.
On FID and IR, QuoRem ranks first in four and five of the eight switched cells, while \ts\ leads IR on SANA-1.6B at $\tau=50\%$ and SD3.5-Large, whereas dual-checkpoint baselines reach slightly lower FID on FLUX.1-dev.
Visually, QuoRem keeps composition and texture up to $\tau\approx 75\%$ (Fig.~\ref{fig:switch_gallery}), and at $\tau=50\%$ it preserves the gold seams, carousel details, spices, and lantern goldfish that \td\ blurs (Fig.~\ref{fig:detail}).

Table~\ref{tab:tmpdiff} and Fig.~\ref{fig:teaser} also explain why prior switching paradigms fall short.
Dual-checkpoint QuaRot and \td\ spend $1.50$ to $1.54\text{ B/w}$ yet lose fidelity to wider symmetric INT4 steps and uncentered token means (Secs.~\ref{sec:quorem} and \ref{sec:mean}).
Sibling stitching in \ts\ ($\text{0.6B}\to\text{1.6B}$ and $\text{Medium}\to\text{Large}$) runs an unquantized smaller BF16 model early, yielding plausible standalone images (hence its FID and IR) while altering object pose and dropping paired PSNR to $10.6$ to $12.5\text{ dB}$ (Fig.~\ref{fig:stitch_align}).

\subsection{System efficiency and Pareto frontier}
\label{sec:systems}\label{sec:frontier}

\begin{table}[!t]
\centering
\caption{\textbf{End-to-end efficiency on one RTX A5000 (24\,GB)} ($1024^2$, $\text{batch}=1$, C++ Nunchaku; \textbf{GiB}: nvidia-smi memory).
\textbf{(a)}~Static; \textbf{(b)}~Switching ($t_{\text{hi}}/t_{\text{lo}}$: W8/W4 step).
$^\ast$Co-resident BF16+INT4 ($29.6\text{ GiB}$) exceeds 24\,GB on FLUX, so TMPDiff swaps over PCIe ($22.58\text{ GiB}$).}
\label{tab:systems}
\scriptsize\setlength{\tabcolsep}{1.7pt}\renewcommand{\arraystretch}{0.88}\setlength{\aboverulesep}{0.08ex}\setlength{\belowrulesep}{0.10ex}\setlength{\ULdepth}{0.9pt}\renewcommand{\ULthickness}{0.34pt}\let\underline\uline
\begin{tabularx}{\columnwidth}{@{}Xlcrrrrrr@{}}
\toprule
\multicolumn{3}{@{}l}{\textbf{(a) Static Execution}} & \multicolumn{3}{c}{\textbf{SANA-1.6B} ($3.12\text{ s}$)} & \multicolumn{3}{c@{}}{\textbf{FLUX.1-dev} ($32.02\text{ s}$)}\\
\cmidrule(lr){4-6}\cmidrule(l){7-9}
\textbf{Method} & \textbf{W/A} & \textbf{B/w}$\downarrow$ & \textbf{GiB}$\downarrow$ & \textbf{ms}$\downarrow$ & $\times$\textbf{BF16}$\uparrow$ & \textbf{GiB}$\downarrow$ & \textbf{ms}$\downarrow$ & $\times$\textbf{BF16}$\uparrow$\\
\midrule
BF16 & 16/16 & 2.00 & 3.86 & 156.0 & 1.00 & 23.81 & 1143.5 & 1.00\\
LLM.int8() & 8/8 & 1.02 & 2.83 & 236.3 & 0.66 & 12.97 & 1481.3 & 0.77\\
TorchAO INT8 & 8/16 & 1.01 & 2.75 & 162.2 & 0.96 & 12.23 & 1331.4 & 0.86\\
Nunchaku W8A8$^\ddagger$ & 8/8 & 1.00 & 2.66 & 64.0 & 2.46 & 10.98 & 641.1 & 1.82\\
NF4 & 4/16 & \textbf{0.56} & \underline{2.40} & 161.9 & 0.96 & \underline{7.63} & 1208.3 & 0.95\\
SVDQuant~\cite{li2025svdquant} & 4/4 & \underline{0.65} & \textbf{2.04} & \textbf{53.2} & \textbf{2.93} & \textbf{7.35} & \textbf{525.6} & \textbf{2.18}\\
\textbf{QuoRem (W8)}\cc & \cc 8/8 & \cc 1.00 & \cc 2.70 & \cc 80.0 & \cc 1.95 & \cc 11.29 & \cc 811.1 & \cc 1.41\\
\textbf{QuoRem (W4)}\cc & \cc 4/4 & \cc 1.00 & \cc 2.70 & \cc\underline{61.6} & \cc\underline{2.53} & \cc 11.29 & \cc\underline{581.9} & \cc\underline{1.97}\\
\bottomrule
\end{tabularx}\\[0.35ex]
\begin{tabularx}{\columnwidth}{@{}Xrrrrrrrr@{}}
\toprule
\multicolumn{4}{@{}l}{\textbf{(b) Temporal Switching}} & \multicolumn{5}{c@{}}{\textbf{Speedup ($\times$BF16) across $\tau$}$\uparrow$}\\
\cmidrule(l){5-9}
\textbf{Method} & \textbf{B/w}$\downarrow$ & \textbf{GiB}$\downarrow$ & $t_{\text{hi}}/t_{\text{lo}}$\textbf{ (ms)}$\downarrow$ & \textbf{0\%} & \textbf{25\%} & \textbf{50\%} & \textbf{75\%} & \textbf{100\%}\\
\midrule
\multicolumn{9}{@{}l}{\textbf{SANA-1.6B} ($20$ steps; BF16 loop: $3.12\text{ s}$)}\\[0.08ex]
T-Stitch$^\dagger$~\cite{pan2024tstitch} & 2.74 & \underline{5.06} & 156.9/76.5 & \underline{1.00} & \underline{1.32} & \underline{1.50} & 1.73 & 2.05\\
TMPDiff~\cite{lewandowski2026tmpdiff} & \underline{2.65} & 5.32 & 156.0/\textbf{53.2} & \underline{1.00} & 1.18 & 1.46 & \underline{1.93} & \textbf{2.93}\\
\textbf{QuoRem}\cc & \cc\textbf{1.00} & \cc\textbf{2.70} & \cc\textbf{80.0}/\underline{61.6} & \cc\textbf{1.95} & \cc\textbf{2.04} & \cc\textbf{2.18} & \cc\textbf{2.34} & \cc\underline{2.53}\\
\midrule
\multicolumn{9}{@{}l}{\textbf{FLUX.1-dev} ($28$ steps; BF16 loop: $32.02\text{ s}$)}\\[0.08ex]
T-Stitch~\cite{pan2024tstitch} & \multicolumn{8}{c@{}}{\textit{N/A (no smaller FLUX.1 sibling architecture)}}\\
TMPDiff$^\ast$~\cite{lewandowski2026tmpdiff} & \underline{2.65} & \underline{22.58}$^\ast$ & 1143.5/\textbf{525.6} & \underline{1.00} & \underline{1.06}$^\ast$ & \underline{1.24}$^\ast$ & \underline{1.50}$^\ast$ & \textbf{2.18}\\
\textbf{QuoRem}\cc & \cc\textbf{1.00} & \cc\textbf{11.29} & \cc\textbf{811.1}/\underline{581.9} & \cc\textbf{1.41} & \cc\textbf{1.51} & \cc\textbf{1.64} & \cc\textbf{1.78} & \cc\underline{1.97}\\
\bottomrule
\end{tabularx}
\end{table}

\paragraph{Memory footprint.}
One $1.00\text{ B/w}$ bank serves both precisions without a second checkpoint, so QuoRem halves TMPDiff's GPU memory on both backbones (Tab.~\ref{tab:systems}) and fits 12B FLUX.1-dev in $11.29\text{ GiB}$ on a 24\,GB RTX A5000, avoiding PCIe swaps where TMPDiff needs $29.6\text{ GiB}$.

\begin{figure}[!t]
\centering
\includegraphics[width=0.94\columnwidth, trim=0 0 0 9pt, clip]{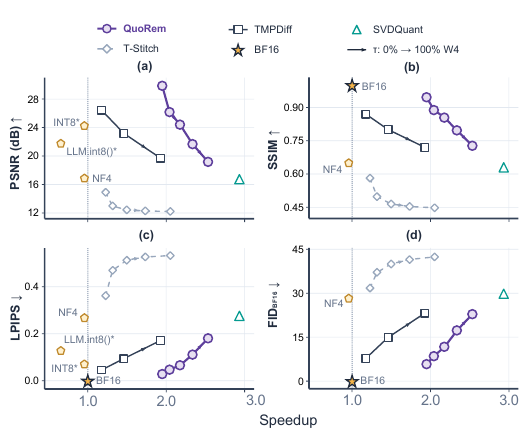}
\caption{\textbf{Speedup--fidelity Pareto frontier on SANA-1.6B} ($1024^2$, $20$ steps, $998$ COCO prompts).
From one $1.00\text{ B/w}$ parent, varying $\tau$ traces a smooth frontier that outperforms dual-model T-Stitch and TMPDiff across all budgets while keeping $2.4$ to $13.1\text{ dB}$ higher PSNR than static SVDQuant.}
\label{fig:trajectory}
\end{figure}

\paragraph{Latency and Pareto frontier.}
With fused $B=64$ rotation and rank-zero \textbf{8-Bit Mean}, QuoRem runs $2.18\times$ faster than BF16 on SANA-1.6B and $1.64\times$ on FLUX.1-dev at $\tau=50\%$, or $1.49\times$ and $1.32\times$ faster than TMPDiff at half its memory (Tab.~\ref{tab:systems}b).
Weight-only NF4 and TorchAO INT8 dequantize weights to BF16 before $1\text{k}$ to $4\text{k}$-token GEMMs instead of running integer Tensor Cores, and LLM.int8() splits outliers into an FP16 GEMM, keeping all three at $0.66$ to $0.96\times$ BF16 (Tab.~\ref{tab:systems}a).
In the same C++ engine, Nunchaku W8A8 beats QuoRem's W8 speed by dropping rotation and $B=64$ group scales at a cost of about $15\text{ dB}$ PSNR on SANA, whereas QuoRem's W4 view from the same $1.00\text{ B/w}$ bank outpaces Nunchaku W8A8 on both backbones ($2.53\times$ vs.\ $2.46\times$ BF16 on SANA).
TMPDiff slows down on BF16 steps and FLUX PCIe swaps, while static SVDQuant is fixed at one point $2.4$ to $13.1\text{ dB}$ below QuoRem's curve on SANA (Fig.~\ref{fig:trajectory}).

\section{Ablation study and mechanism analysis}
\label{sec:analysis}

\subsection{Parent alignment and component ablations}
\label{sec:deletion}

\begin{table}[!t]
\centering
\caption{\textbf{4-bit weight decomposition ablation} on all $1{,}074$ quantized linear layers of the four DiTs ($\text{NMSE}_W=\sum_\ell\|W_\ell-\W_\ell\|_F^2/\sum_\ell\|W_\ell\|_F^2\downarrow$, rotated weights).}
\label{tab:parent_ablation}\label{tab:policy}
\scriptsize\setlength{\tabcolsep}{3.0pt}\renewcommand{\arraystretch}{0.92}\setlength{\aboverulesep}{0.10ex}\setlength{\belowrulesep}{0.12ex}
\begin{tabularx}{\columnwidth}{@{}Xcrrrr@{}}
\toprule
\textbf{Weight Decomposition} & \textbf{B/w}$\downarrow$ & \textbf{SANA-1.6B} & \textbf{SD3.5-M} & \textbf{SD3.5-L} & \textbf{FLUX.1-dev}\\
\midrule
Floor Slicing & \textbf{1.00} & 0.0654 & 0.0664 & 0.0707 & 0.0766\\
Symmetric INT4 & \underline{1.50} & \underline{0.0233} & \underline{0.0235} & \underline{0.0250} & \underline{0.0271}\\
\textbf{Offset Q--R (Ours)}\cc & \cc\textbf{1.00} & \cc\textbf{0.0182} & \cc\textbf{0.0185} & \cc\textbf{0.0197} & \cc\textbf{0.0213}\\
\bottomrule
\end{tabularx}
\end{table}

\begin{figure*}[!t]
\centering
\includegraphics[width=\textwidth]{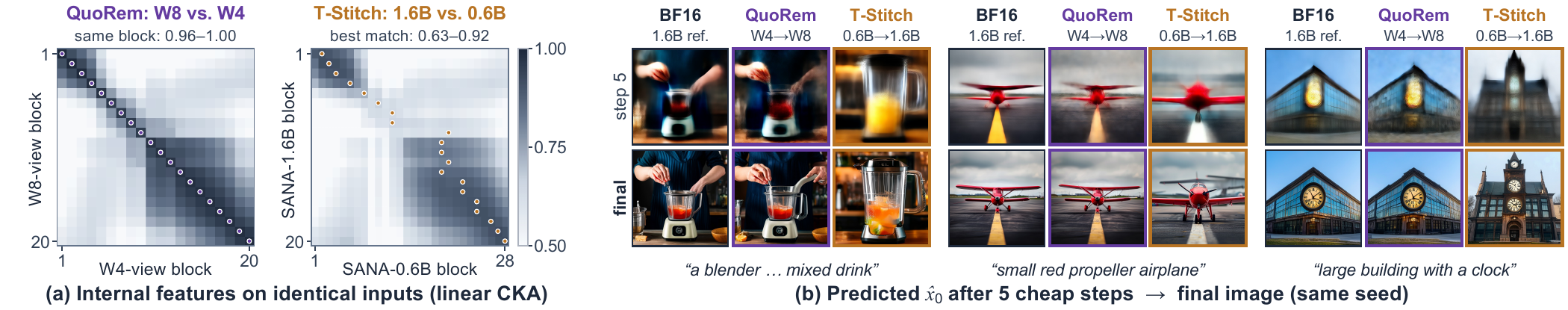}
\caption{\textbf{Feature alignment: single-parent switching vs.\ sibling stitching on SANA-1.6B ($1024^2$).}
\textbf{(a)}~Blockwise linear CKA (32 prompts $\times$ 20 steps, RTX A5000; dots: best match): QuoRem's W8/W4 views stay aligned ($\text{CKA}=0.96\text{--}1.00$), whereas middle SANA-1.6B blocks drop to $0.63$ vs.\ SANA-0.6B.
\textbf{(b)}~Step-5 $\hat{x}_0$ and final output: QuoRem preserves BF16 layout altered by SANA-0.6B.}
\label{fig:stitch_align}
\end{figure*}

\paragraph{Single-parent views vs.\ model stitching.}
QuoRem derives its W8 and W4 views from the same weights so their internal features match closely ($\text{CKA}=0.96$ to $1.00$), whereas sibling stitching (\ts~\cite{pan2024tstitch}) hands features from SANA-0.6B into middle SANA-1.6B blocks whose linear CKA~\cite{kornblith2019cka} drops to $0.63$ on identical inputs (Fig.~\ref{fig:stitch_align}a), refining a shifted layout and changing object pose (Fig.~\ref{fig:stitch_align}b).

\paragraph{Offset Q-R decomposition.}
Across all $1{,}074$ layers, $+8$ Offset Q-R (Sec.~\ref{sec:quorem}, Tab.~\ref{tab:parent_ablation}) cuts 4-bit $\text{NMSE}_W$ by $3.59\times$ over Floor Slicing and by $21.2$ to $21.7\%$ over $1.50\text{ B/w}$ Symmetric INT4 with an exact W8 view at $1.00\text{ B/w}$.
End to end, replacing floor slicing with $+8$ Offset Q-R raises all-W4 FLUX PSNR by $2.02\text{ dB}$ and cuts all-W4 SD3.5-L $\text{FID}_{\text{BF16}}$ by $70.3$ points, beating $1.50\text{ B/w}$ QuaRot$^\dagger$ (Fig.~\ref{fig:teaser}).

\paragraph{Cross-plane mean restoration and $r=32$ control.}
Block-Hadamard rotation lowers SANA's peak-to-RMS outlier ratio from $17.0$ to $4.8$, yet $52.5\%$ of activation energy stays in $\mu$.
Restoring $\mu\W_8^\top$ with 8-Bit Mean (Sec.~\ref{sec:mean}, Tab.~\ref{tab:mean_ablation}) cancels $\one\mu\widehat R_4^\top$ and lowers $\text{MSE}_v$ by $1.52$ to $3.21\times$ over Uncentered ($1.29\times$ on FLUX at matched $\alpha=0.8$), whereas 4-Bit Mean ($\mu\W_4^\top$) keeps the remainder bias of Eq.~\eqref{eq:meanerror} and incurs $10$ to $30\%$ higher $\text{MSE}_v$.
To isolate online centering from the uncentered +\,Offset Q-R baseline in Fig.~\ref{fig:teaser} (which uses $\alpha=0.9$ without Static FF-Shift), adding 8-Bit Mean alone on COCO-998 raises all-W4 PSNR by $0.69$ to $3.65\text{ dB}$ across all four DiTs ($+3.65\text{ dB}$ on SANA) and cuts $\tau=50\%$ $\text{FID}_{\text{BF16}}$ on SD3.5-M/L by $12.86$ and $7.38$ points.
On SD3.5-M/L, +\,Static FF-Shift stays within $1.06\times$ the $\text{MSE}_v$ of 8-Bit Mean, removes online FF-out reductions, and adds $0.11$ and $0.32\text{ dB}$ PSNR.
Because $\mathbf{R}$ already stores $\widehat W_8-\widehat W_4$ and 8-Bit Mean cancels its shared-mean error, adding a rank-32 BF16 adapter duplicates $\widehat R_4$, exceeds the $1.00\text{ B/w}$ parent, and gains only $0.14$ to $0.26\text{ dB}$ all-W4 PSNR while slowing all-W4 execution by $13.2$ to $49.8\%$.

\begin{table}[!t]
\centering
\caption{\textbf{4-bit mean restoration ablation.}
Velocity error $\text{MSE}_v\downarrow$ against BF16 on held-out trajectory states; shaded cells mark the deployed policy selected by our $\le 1.10\times$ $\text{MSE}_v$ and 16-prompt quality gates.}
\label{tab:mean_ablation}
\scriptsize\setlength{\tabcolsep}{4.8pt}\renewcommand{\arraystretch}{0.94}\setlength{\aboverulesep}{0.09ex}\setlength{\belowrulesep}{0.11ex}
\begin{tabularx}{\columnwidth}{@{}Xrrrr@{}}
\toprule
\textbf{Mean Restoration} & \textbf{SANA-1.6B} & \textbf{SD3.5-M} & \textbf{SD3.5-L} & \textbf{FLUX.1-dev}\\
\midrule
Uncentered & 0.0212 & 0.0322 & 0.0211 & 0.0288\\
4-Bit Mean ($\mu\W_4^\top$) & 0.0086 & 0.0222 & 0.0162 & 0.0209\\
\textbf{8-Bit Mean ($\mu\W_8^\top$)}\rule{0pt}{2.45ex} & \cc\textbf{0.0066} & \textbf{0.0183} & \textbf{0.0136} & \cc\textbf{0.0189}\\
\textbf{+ Static FF-Shift} & \underline{0.0068} & \cc\underline{0.0193} & \cc\underline{0.0141} & \underline{0.0199}\\
\bottomrule
\end{tabularx}
\end{table}

\begin{table}[!t]
\centering
\caption{\textbf{Schedule comparison at $\tau=50\%$} (controlled track; 256 prompts on SD3.5/FLUX, 998 on SANA; $a/b/c(/d)$: W8/W4 phase lengths).}
\label{tab:schedule_comparison}
\scriptsize\setlength{\tabcolsep}{2.1pt}\renewcommand{\arraystretch}{0.88}\setlength{\aboverulesep}{0.08ex}\setlength{\belowrulesep}{0.10ex}
\begin{tabularx}{\columnwidth}{@{}Xrrrrrr@{}}
\toprule
\textbf{Schedule} & \textbf{PSNR}$\uparrow$ & \textbf{SSIM}$\uparrow$ & \textbf{LPIPS}$\downarrow$ & \textbf{FID}$_{\text{BF16}}\downarrow$ & \textbf{FID}$\downarrow$ & \textbf{IR}$\uparrow$\\
\midrule
\multicolumn{7}{@{}l}{\textbf{SD3.5-Medium} ($512^2$, $N=28$, $k=14$ W4 steps)}\\
Ascend & 14.14 & 0.597 & 0.411 & \underline{75.14} & \textbf{134.81} & \textbf{1.052}\\
Descend & \underline{19.57} & \underline{0.719} & \underline{0.233} & 78.07 & 139.32 & 0.967\\
\textbf{Waterline (12/13/2/1)}\cc & \cc\textbf{19.66} & \cc\textbf{0.764} & \cc\textbf{0.202} & \cc\textbf{56.27} & \cc\underline{135.67} & \cc\underline{1.022}\\
\midrule
\multicolumn{7}{@{}l}{\textbf{SD3.5-Large} ($1024^2$, $N=28$, $k=14$ W4 steps)}\\
Ascend & 13.09 & 0.593 & 0.478 & 84.21 & \textbf{133.49} & \textbf{1.122}\\
Descend & \underline{18.70} & \underline{0.748} & \underline{0.263} & \underline{67.90} & 137.97 & 1.093\\
\textbf{Waterline (12/13/2/1)}\cc & \cc\textbf{18.80} & \cc\textbf{0.782} & \cc\textbf{0.239} & \cc\textbf{58.19} & \cc\underline{134.95} & \cc\underline{1.095}\\
\midrule
\multicolumn{7}{@{}l}{\textbf{SANA-1.6B} ($1024^2$, $N=20$, $k=10$ W4 steps)}\\
Ascend & \underline{17.79} & \underline{0.691} & \underline{0.229} & \underline{26.11} & \underline{71.08} & \textbf{1.033}\\
\textbf{Descend (= Waterline)}\cc & \cc\textbf{21.96} & \cc\textbf{0.784} & \cc\textbf{0.107} & \cc\textbf{16.75} & \cc\textbf{71.06} & \cc\underline{1.015}\\
\midrule
\multicolumn{7}{@{}l}{\textbf{FLUX.1-dev} ($1024^2$, $N=28$, $k=14$ W4 steps)}\\
Ascend & \underline{18.67} & \underline{0.744} & \underline{0.301} & \underline{57.16} & \textbf{135.73} & \underline{1.027}\\
\textbf{Descend (= Waterline)}\cc & \cc\textbf{26.72} & \cc\textbf{0.901} & \cc\textbf{0.093} & \cc\textbf{22.74} & \cc\underline{136.25} & \cc\textbf{1.046}\\
\bottomrule
\end{tabularx}
\end{table}

\subsection{Waterline mechanism analysis and scheduling}
\label{sec:exp_direction}
\begin{figure}[!t]
\centering
\includegraphics[width=\columnwidth]{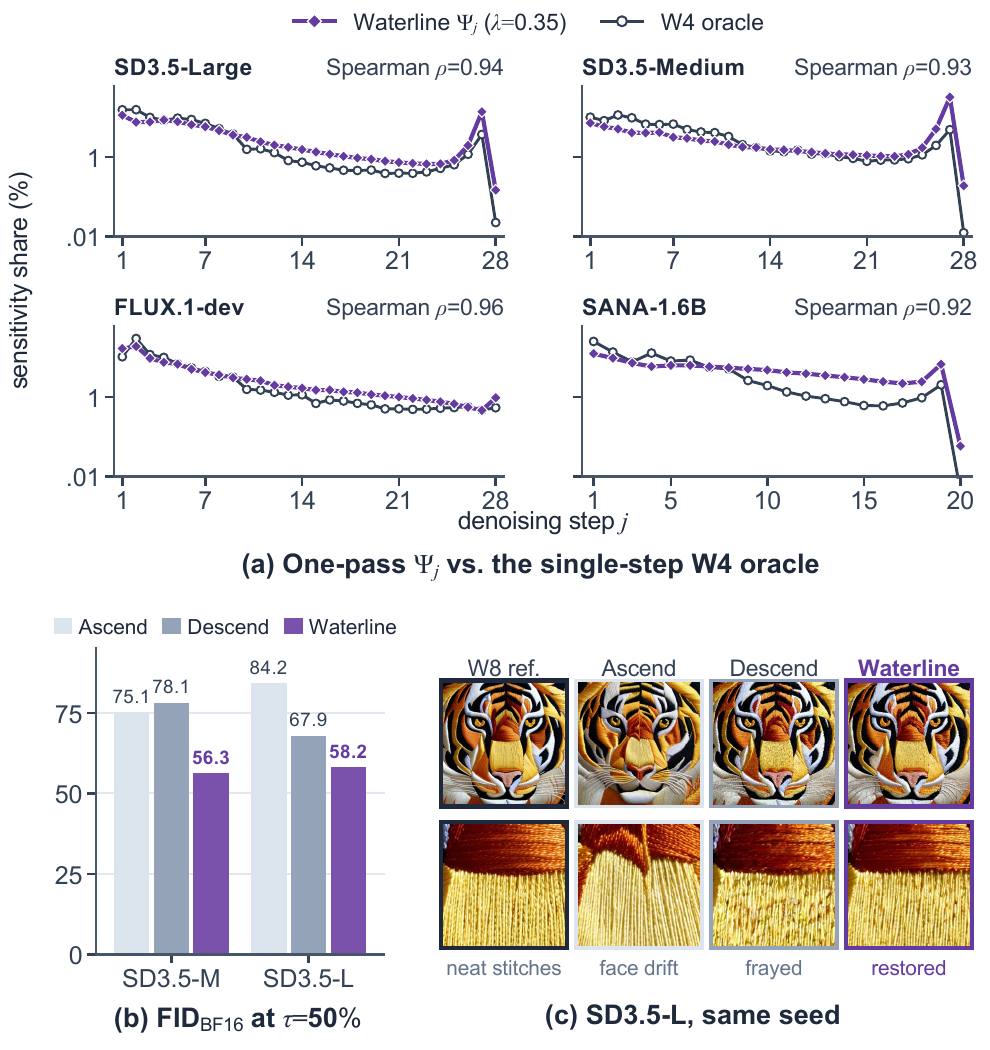}
\caption{\textbf{Waterline validation across DiTs.}
\textbf{(a)}~One-pass $\Psi_j$ tracks the $N$-rollout oracle (log; $\rho=0.92\text{--}0.96$) and captures the step-27 spike that SD3.5 shows and SANA/FLUX lack.
\textbf{(b)}~$\mathrm{FID}_{\mathrm{BF16}}$ ($\tau=50\%$).
\textbf{(c)}~SD3.5-L: \textbf{Waterline} avoids face drift and frayed stitches (visual cases in Appendix~\ref{sec:supp_visuals}).}
\label{fig:waterline_valid}
\end{figure}

\paragraph{Accuracy and stability of the one-pass profiler.}
Measuring the single-step W4 oracle takes $N$ rollouts per prompt, whereas our shadow profiler runs the resident W4 view along $4$ calibration prompts in $<6\text{ min}$ on RTX A5000.
Across all four DiTs (Fig.~\ref{fig:waterline_valid}a), $\Psi_j(\lambda)$ ranks the oracle's whitened terminal error with Spearman $\rho=0.92$ to $0.96$ ($0.89$ to $0.98$ on held-out C++ SANA/FLUX profiles).
Sweeping $\lambda\in[0.20,0.50]$ or $\varepsilon_0\in[0.05,0.12]$ around $(0.35,0.08)$ shifts $S^\star(k)$ by $\le 1$ interior step and keeps every regime choice.

\paragraph{How mean restoration shapes the schedule across models.}
Cross-backbone $\Psi_j(\lambda)$ differences (Fig.~\ref{fig:waterline_valid}a, Tab.~\ref{tab:schedule_comparison}) reflect how mean restoration (Sec.~\ref{sec:mean}) controls tail error $d_j$.
On SANA-1.6B and FLUX.1-dev, online 8-Bit Mean removes $\mu$ per step to keep tail error low, so $\Delta_{\text{WL}}<\delta_{\min}$ and Waterline selects Descend.
On SD3.5-M/L (CFG $7.0$), the shifted flow schedule takes a large late jump at step $27$ that clears $91.9\%$ of the remaining noise, where offline Static FF-Shift leaves high relative velocity error ($\mathcal{E}_{27}^{\text{rel}}=66.5\%$ to $134.2\%$).
This tail spike lifts $\Psi_{26}$ and $\Psi_{27}$ above the $\tau=50\%$ cutoff $\Psi_{(14)}$ so both stay protected in W8, whereas the $11.5\times$ smaller final step $28$ ($\Delta t_{28}=0.0089$) remains safely below the threshold.

\begin{table}[!t]
\centering
\caption{\textbf{PlaneFold switching overhead} on RTX A5000.
\textbf{Swap}: max slot ($57.4/72.0\text{ MiB}$ pool); $\ell(0)/\ell(N)$: W8/W4 loop; \textbf{Fold}: 2-switch round-trip (\% of $\tau=50\%$); \textbf{Lin.\ Res.}: max linear residual; \textbf{Exact}: byte-exact.}
\label{tab:handoff}
\scriptsize\setlength{\tabcolsep}{3.2pt}\renewcommand{\arraystretch}{0.92}\setlength{\aboverulesep}{0.08ex}\setlength{\belowrulesep}{0.10ex}
\begin{tabularx}{\columnwidth}{@{}Xrrrrr@{}}
\toprule
\textbf{Backbone} & \textbf{Swap} & $\ell(0)/\ell(N)$\textbf{ (s)}$\downarrow$ & \textbf{Fold (ms)}$\downarrow$ & \textbf{Lin.\ Res.}$\downarrow$ & \textbf{Exact}\\
\midrule
SANA-1.6B & 24.8\,MiB & 1.63 / 1.25 & 8.9 (0.61\%) & 1.24\% & 120/120\\
FLUX.1-dev & 36.0\,MiB & 22.98 / 16.52 & 51.8 (0.26\%) & 0.14\% & 304/304\\
\bottomrule
\end{tabularx}
\end{table}

\paragraph{Comparing schedules and reclaiming the final step.}
At $\tau=50\%$ (Fig.~\ref{fig:waterline_valid}b, Tab.~\ref{tab:schedule_comparison}), Ascend's late W8 half refines textures (holding FID/IR), whereas early W4 drift redraws the tiger's face (SD3.5-L, Fig.~\ref{fig:waterline_valid}c).
Descend keeps early layout and beats Ascend by $5.4$ to $5.6\text{ dB}$ PSNR on SD3.5-M/L, yet leaving steps $26$ and $27$ in W4 frays the stitches.
Waterline ($12/13/2/1$ alternating W8/W4 phases, i.e., $S^\star(14)=\{13,\ldots,25,28\}$) protects both the layout transition at step $12$ and the late jump at steps $26$--$27$ in W8 by reclaiming tiny step $28$ for W4, winning every paired metric on SD3.5-M/L and cutting SD3.5-M $\text{FID}_{\text{BF16}}$ from $78.07$ to $56.27$.

\subsection{Switching overhead and handoff analysis}
\label{sec:exp_handoff}
\paragraph{Switching overhead and linear latency law.}
PlaneFold (Sec.~\ref{sec:runtime}, Tab.~\ref{tab:handoff}) replaces a $2$ to $11\%$ per-step $(\mathbf{Q},\mathbf{R})$ unpack on W8 steps with one fold per switch via one swap slot, so a W8$\to$W4$\to$W8 round trip costs $0.61\%$ of the $\tau=50\%$ loop on SANA and $0.26\%$ on FLUX.
Measured $\ell(\pi)$ follows the linear law within $1.24\%$ at every $k\in[0,N]$, and $H(\pi)$ stays below $1.2\%$ even at four switches.

\paragraph{Bit-exact and smooth precision transitions.}
All $120/120$ $(\mathbf{Q},\mathbf{R})$ layers on SANA and $304/304$ on FLUX reconstruct losslessly across $(\mathbf{Q},\mathbf{R})\to\mathbf{Z}\to(\mathbf{Q},\mathbf{R})$, and $\text{W4}\to\text{W8}\to\text{W4}$ matches unswitched W4 bit-for-bit.
Because both W8 and W4 views operate in the exact same Hadamard-rotated coordinate frame and share $\one(\mu\W_8^\top)$, switching introduces no representation mismatch or sudden shifts in channel averages; moving from one-switch Descend to three-switch Waterline (Tab.~\ref{tab:schedule_comparison}) thus improves PSNR on $155/256$ SD3.5-M and $184/256$ SD3.5-L prompts.

\section{Conclusion}
\label{sec:conclusion}
We presented QuoRem, a single-parent framework for temporal precision switching in DiTs that runs W8A8 and adapter-free ($r=0$) W4A4 from $1.00\text{ B/w}$ weights.
Offline offset quotient-remainder coding and online cross-plane mean restoration remove 4-bit slicing bias and shared token-mean drift with zero low-rank branches.
One-pass Waterline profiling models early layout drift and late detail exposure to unify one-switch and multi-phase schedules, while PlaneFold folds planes in place at switches with minimal overhead.
Across four 1.6B--12B DiTs, QuoRem halves dual-checkpoint GPU memory, delivers up to $2.18\times$ speedup over BF16, and sets the strongest speed--fidelity Pareto frontier.

{\small\bibliographystyle{ieeenat_fullname}\bibliography{apd_refs}}

\clearpage
\maketitlesupplementary
\appendix
\setcounter{equation}{0}
\setcounter{table}{0}
\setcounter{figure}{0}
\setcounter{theorem}{0}
\setcounter{proposition}{0}
\setcounter{lemma}{0}
\renewcommand{\theequation}{S\arabic{equation}}
\renewcommand{\thetable}{S\arabic{table}}
\renewcommand{\thefigure}{S\arabic{figure}}
\renewcommand{\theHequation}{S\arabic{equation}}
\renewcommand{\theHtable}{S\arabic{table}}
\renewcommand{\theHfigure}{S\arabic{figure}}

\section{Proofs of Formal Claims}
\label{sec:supp_proofs}

\subsection{Offset \texorpdfstring{$+8$}{+8} Quotient--Remainder Decomposition}
\label{sec:supp_quorem}

For a signed INT8 parent code $z\in\mathcal{Z}=\{-128,\dots,119\}$ ($|\mathcal{Z}|=248$) with step scale $s>0$, Sec.~\ref{sec:quorem} and Eq.~\eqref{eq:quorem} define $q_4=\lfloor(z+8)/16\rfloor$, $r_4=z+8-16q_4$, and $\W_8=\W_4+\widehat R_4$ with $\W_8=s\mathbf{Z}$, $\W_4=16s\mathbf{Q}$, and $\widehat R_4=s(\mathbf{R}-8)$.

\begin{theorem}[Offset quotient--remainder decomposition]
\label{thm:quorem}
Let $\mathcal{Z}=\{-128,\dots,119\}$ and $\Delta_4=16s$. Then:
\begin{enumerate}
\item \textbf{Bijection vs.\ RNE tie overflow.} Equation~\eqref{eq:quorem} maps $\mathcal{Z}$ injectively into $(q_4,r_4)\in[-8,7]\times[0,15]$ with exact inverse $z=16q_4+r_4-8$, whereas round-to-nearest-even $q_{\mathrm{rne}}=\operatorname{RNE}(z/16)$ maps $17$ integer codes $\{16q-8,\dots,16q+8\}$ to each interior even quotient $q$, overflowing a $16$-state 4-bit remainder.
\item \textbf{Bias and MSE optimality.} On an interior cell with uniform discrete residue $U\sim\operatorname{Unif}\{0,\dots,15\}$ and independent centered parent residual $\eta\sim\operatorname{Unif}[-1/2,1/2]$ (in units of $s$), $q_4=\lfloor(z+c)/16\rfloor$ ($c\in\mathbb{Z}$) gives bias $\E[16sq_4-sz]=(c-7.5)s$ and MSE $\E[(16sq_4-w)^2]=\big(\tfrac{256}{12}+(c-7.5)^2\big)s^2$. Offsets $c\in\{7,8\}$ minimize both over $c\in\mathbb{Z}$: $c=8$ reduces $|\text{bias}|$ by $15\times$ vs.\ floor slicing ($c=0$, from $-7.5s$ to $+0.5s$) and lowers MSE by $931/259\approx 3.59\times$:
\begin{equation}
\frac{259}{256}\frac{\Delta_4^2}{12}\;\;(c=8)
\quad\text{vs.}\quad
\frac{931}{256}\frac{\Delta_4^2}{12}\;\;(c=0).
\label{eq:supp_nesting}
\end{equation}
\item \textbf{Comparison to symmetric INT4.} Relative to $15$-level symmetric INT4 ($q_{\mathrm{sym}}\in[-7,7]$, step $s_{\mathrm{sym}}=127s/7$), using all $16$ levels $q_4\in[-8,7]$ at step $16s$ achieves granular MSE ratio $\frac{259\cdot 7^2}{127^2}=\frac{12691}{16129}\approx 0.7868$ ($21.32\%$ reduction).
\end{enumerate}
\end{theorem}
\begin{proof}
(1) Since $z+8\in[-120,127]$, Euclidean division by $16$ yields a unique $q_4\in[-8,7]$ and $r_4\in[0,15]$ ($8+15\cdot 16=248$ codes), inverted by $z=16q_4+r_4-8$. Under $\operatorname{RNE}(z/16)$, both ties $16q\pm 8$ round to even $q$, giving $17>2^4$ inputs.
(2) Here $16q_4-z=c-U$ has mean $c-7.5$ and variance $255/12$; adding $\operatorname{Var}(\eta)=1/12$ gives $\big(\tfrac{256}{12}+(c-7.5)^2\big)s^2$, minimized over $\mathbb{Z}$ at $c\in\{7,8\}$ ($259s^2/12$ at $c=8$ vs.\ $931s^2/12$ at $c=0$).
(3) Dividing $259s^2/12$ by $s_{\mathrm{sym}}^2/12=(127/7)^2s^2/12$ gives $12691/16129$.
\end{proof}

\subsection{Rank-Zero Cross-Plane Mean Restoration}
\label{sec:supp_mean}

In Sec.~\ref{sec:mean}, for rotated activations $X=X_{\mathrm{raw}}\widetilde H_B\in\R^{T\times D}$ under orthogonal $\widetilde H_B=I_{D/B}\otimes H_B$, the token mean $\mu=T^{-1}\one^\top X=(T^{-1}\one^\top X_{\mathrm{raw}})\widetilde H_B\in\R^{1\times D}$ ($\one\in\R^{T\times 1}$) satisfies $\|\mu\|_2=\|T^{-1}\one^\top X_{\mathrm{raw}}\|_2$. With $X_c=X-\one\mu$, Eq.~\eqref{eq:m8} evaluates $Y_{\text{online}}=\mathcal{Q}_4(X_c)\W_4^\top+\one(\mu\W_8^\top)+b$.

\begin{proposition}[Rank-zero shared-mean cancellation]
\label{prop:mean}
With $\W_8=\W_4+\widehat R_4$ and $E_4=\mathcal{Q}_4(X_c)-X_c$, Eq.~\eqref{eq:meanerror} holds in real arithmetic,
\[
Y_{\text{online}}-(X\W_8^\top+b)=E_4\W_4^\top-X_c\widehat R_4^\top,
\]
where $\one^\top(X_c\widehat R_4^\top)=0$ and $\mu\W_8^\top\in\R^{1\times D_{\mathrm{out}}}$ costs one GEMV ($O(1/T)$ of the GEMM) at $r=0$, whereas $\one(\mu\W_4^\top)$ leaves bias $-\one(\mu\widehat R_4^\top)$. For Static FF-Shift ($b_t'=b+\bar\mu_t\W_8^\top$, $Y_{\text{static},t}=\mathcal{Q}_4(X_t-\one\bar\mu_t)\W_4^\top+b_t'$), with $E_4'(U)=\mathcal{Q}_4(U)-U$,
\begin{equation}
Y_{\text{static},t}-Y_{\text{online}}
=\one(\bar\mu_t-\mu_t)\widehat R_4^\top
+\big[E_4'(X_t-\one\bar\mu_t)-E_4\big]\W_4^\top.
\label{eq:supp_ffshift}
\end{equation}
\end{proposition}
\begin{proof}
Substituting $\mathcal{Q}_4(X_c)=X_c+E_4$ and $X=X_c+\one\mu$ into Eq.~\eqref{eq:m8} yields $(X_c+E_4)\W_4^\top+\one\mu\W_8^\top-(X_c+\one\mu)\W_8^\top=E_4\W_4^\top-X_c\widehat R_4^\top$, with $\one^\top X_c=0$. Using $\one\mu\W_4^\top$ adds $-\one\mu\widehat R_4^\top$, and expanding $Y_{\text{static},t}-Y_{\text{online}}$ gives Eq.~\eqref{eq:supp_ffshift}.
\end{proof}

\subsection{Euler--Wiener Waterline Scheduling}
\label{sec:supp_waterline}

In Sec.~\ref{sec:scheduling}, along $t_0=1>t_1>\dots>t_N=0$ ($\Delta t_j=t_{j-1}-t_j>0$, $N\ge 2$), $z_t=(1-t)\hat{x}_0(z_t,t)+t\hat{\varepsilon}(z_t,t)$ and $v=\hat{\varepsilon}-\hat{x}_0$, so the Euler step $z_r=z_{r-1}-\Delta t_r v(z_{r-1},t_{r-1})$ equals $(1-t_r)\hat{x}_0(z_{r-1},t_{r-1})+t_r\hat{\varepsilon}(z_{r-1},t_{r-1})$.

\begin{lemma}[Euler--Wiener survival gain bound]
\label{lem:ew_gain}
For $x_0\sim\mathcal{N}(\mu_f,\sigma_f^2)$ ($\sigma_f^2>0$) with MMSE Wiener denoiser $\hat{x}_0(z_t)=\E[x_0\mid z_t]$ and $s_f(t)=(1-t)^2\sigma_f^2+t^2$, the step-$r$ Jacobian is
\begin{align}
\phi_{r,f}
&=\frac{(1-t_r)(1-t_{r-1})\sigma_f^2+t_r t_{r-1}}{s_f(t_{r-1})}\nonumber\\
&=1-\Delta t_r\frac{t_{r-1}-(1-t_{r-1})\sigma_f^2}{s_f(t_{r-1})}>0,
\label{eq:supp_phi}
\end{align}
and $\kappa_{j,f}^{\text{E}}=\prod_{r=j+1}^N\phi_{r,f}$ satisfies Eq.~\eqref{eq:ew}:
\[
0<\kappa_{j,f}^{\text{E}}\le\sqrt{\frac{\sigma_f^2}{(1-t_j)^2\sigma_f^2+t_j^2}}.
\]
\end{lemma}
\begin{proof}
Substituting $\partial_z\hat{x}_0=(1-t)\sigma_f^2/s_f(t)$ and $\partial_z\hat{\varepsilon}=t/s_f(t)$ into $\partial_{z_{r-1}}z_r$ gives Eq.~\eqref{eq:supp_phi}. Writing $u_f(t)=\big((1-t)\sqrt{\sigma_f^2},\,t\big)\in\R_{\ge 0}^2$ with angle $\theta_f(t)=\arctan\!\big(\frac{t}{(1-t)\sqrt{\sigma_f^2}}\big)$ yields $\phi_{r,f}=\sqrt{s_f(t_r)/s_f(t_{r-1})}\cos\Delta\theta_{r,f}$ with $\cos\Delta\theta_{r,f}\in(0,1]$; telescoping from $r=j+1$ to $N$ ($s_f(0)=\sigma_f^2$) proves Eq.~\eqref{eq:ew}.
\end{proof}

\begin{theorem}[Waterline sublevel-set selection]
\label{thm:direction}
For $\Psi_j(\lambda)=a_j+\lambda d_j\ge 0$ in Eq.~\eqref{eq:direction}, let $(j_1,\dots,j_N)$ sort $(\Psi_j(\lambda),j)$ lexicographically and set $\Psi_{(k)}=\Psi_{j_k}(\lambda)$. Under $\mathcal{J}_\lambda(S)=\sum_{j\in S}\Psi_j(\lambda)$ over $|S|=k$:
\begin{enumerate}
\item \textbf{Optimality and nesting.} $S^\star(k)=\{j_1,\dots,j_k\}$ in Eq.~\eqref{eq:waterline} minimizes $\mathcal{J}_\lambda(S)$ over $|S|=k$ and satisfies $S^\star(1)\subset\dots\subset S^\star(N)$.
\item \textbf{Final-step reclamation.} Since $a_N\propto\Delta t_N$ and $d_N\propto(\Delta t_N)^2$, if $\Delta t_N\ll\Delta t_{N-1}$ gives $\Psi_N(\lambda)<\Psi_{(k)}<\min\big(\Psi_{h+1}(\lambda),\Psi_{N-1}(\lambda)\big)$, then $N\in S^\star(k)$ and $N-1\notin S^\star(k)$, lowering $\mathcal{J}_\lambda$ by $\Psi_{h+1}(\lambda)-\Psi_N(\lambda)>0$ relative to $\{h+1,\dots,h+k\}$.
\end{enumerate}
\end{theorem}
\begin{proof}
(1) Any $S\ne S^\star(k)$ of size $k$ replaces $m=|S\setminus S^\star(k)|\ge 1$ elements of score $\le\Psi_{(k)}$ with $m$ elements of score $\ge\Psi_{(k+1)}$, so $\mathcal{J}_\lambda(S)-\mathcal{J}_\lambda(S^\star(k))\ge m(\Psi_{(k+1)}-\Psi_{(k)})\ge 0$, and prefixes of $(j_1,\dots,j_N)$ nest.
(2) Swapping $h+1$ for $N$ preserves $|S|=k$, keeps $N-1$ in W8, and reduces $\mathcal{J}_\lambda$ by $\Psi_{h+1}(\lambda)-\Psi_N(\lambda)>0$.
\end{proof}

\subsection{In-Place PlaneFold and Linear Loop Latency}
\label{sec:supp_planefold}

\begin{proposition}[PlaneFold and linear loop latency]
\label{prop:frontier}
In Sec.~\ref{sec:runtime}, let $\pi\in\{\text{W8A8},\text{W4A4}\}^N$ run $k$ W4 steps with $n_{8\to 4}$ and $n_{4\to 8}$ precision switches:
\begin{enumerate}
\item \textbf{Storage bound and lossless folding.} Any fixed-width scalar code for all $248$ values in $\mathcal{Z}$ requires $\lceil\log_2 248\rceil=8$ bits ($1.00\text{ B/w}$). On an even-length weight buffer of $M_w$ coefficients, $[\mathbf{Q};\mathbf{R}]$ occupies $\tfrac{1}{2}M_w+\tfrac{1}{2}M_w=M_w$ bytes, matching $\mathbf{Z}$, and folding $z=16q_4+r_4-8$ vs.\ unfolding $q_4=\lfloor(z+8)/16\rfloor$, $r_4=z+8-16q_4$ via one single-layer swap slot of $\max_\ell M_{w,\ell}$ bytes is an exact byte involution on $\mathcal{Z}^{M_w}$ with zero host--device weight transfer.
\item \textbf{Linear loop latency.} With step costs $t_{\mathrm{hi}}>t_{\mathrm{lo}}$, directional fold costs $h_{8\to 4},h_{4\to 8}$, and fixed intercept $H_0$, trajectory latency obeys Eq.~\eqref{eq:latency}: $\ell(\pi)=(N-k)t_{\mathrm{hi}}+kt_{\mathrm{lo}}+H(\pi)$ with $H(\pi)=n_{8\to 4}h_{8\to 4}+n_{4\to 8}h_{4\to 8}+H_0$.
\end{enumerate}
\end{proposition}
\begin{proof}
(1) Since $2^7<248\le 2^8$, fixed-width storage needs $\ge 8$ bits ($1.00\text{ B/w}$). By Theorem~\ref{thm:quorem}(1), Eq.~\eqref{eq:quorem} is bijective on $\mathcal{Z}$, and staging each layer through the reusable $\max_\ell M_{w,\ell}$ device slot permutes layout in place without PCIe traffic.
(2) Summing $N-k$ W8 steps, $k$ W4 steps, $n_{8\to 4}$ folds, $n_{4\to 8}$ unfolds, and $H_0$ yields Eq.~\eqref{eq:latency}.
\end{proof}

\section{Additional Visual Examples}
\label{sec:supp_visuals}

\begin{figure*}[!p]
\centering
\includegraphics[width=\textwidth]{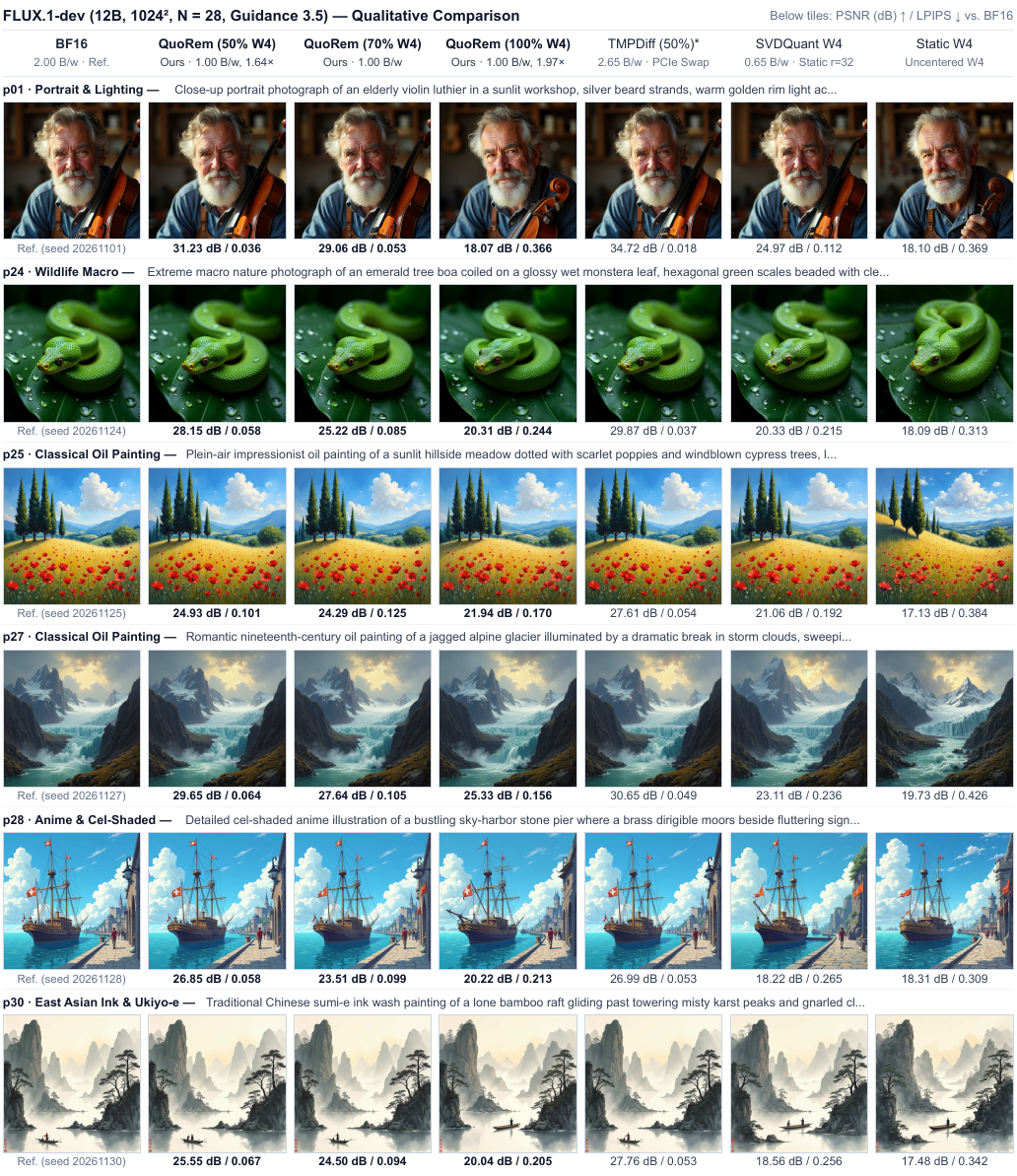}
\caption{\textbf{Full-page $1024\times 1024$ qualitative gallery on FLUX.1-dev} ($12$B, $N=28$, guidance $3.5$; \ts\ is inapplicable as FLUX.1 has no smaller sibling model, cf.\ Tab.~\ref{tab:systems}b). Single-parent \textbf{QuoRem} ($r=0$, $1.00\text{ B/w}$, $11.29\text{ GiB}$) preserves fine portrait hair, macro scale bead reflections, impressionist brushwork, cel-shaded rigging, and sumi-e ink washes at $\tau=50\%$ ($1.64\times$ BF16 speed) and $\tau=70\%$ without dual-checkpoint PCIe swapping (\td$^\ast$, $2.65\text{ B/w}$) or static 4-bit detail loss (SVDQuant W4, Static W4).}
\label{fig:supp_s1}
\end{figure*}

\begin{figure*}[!p]
\centering
\includegraphics[width=\textwidth]{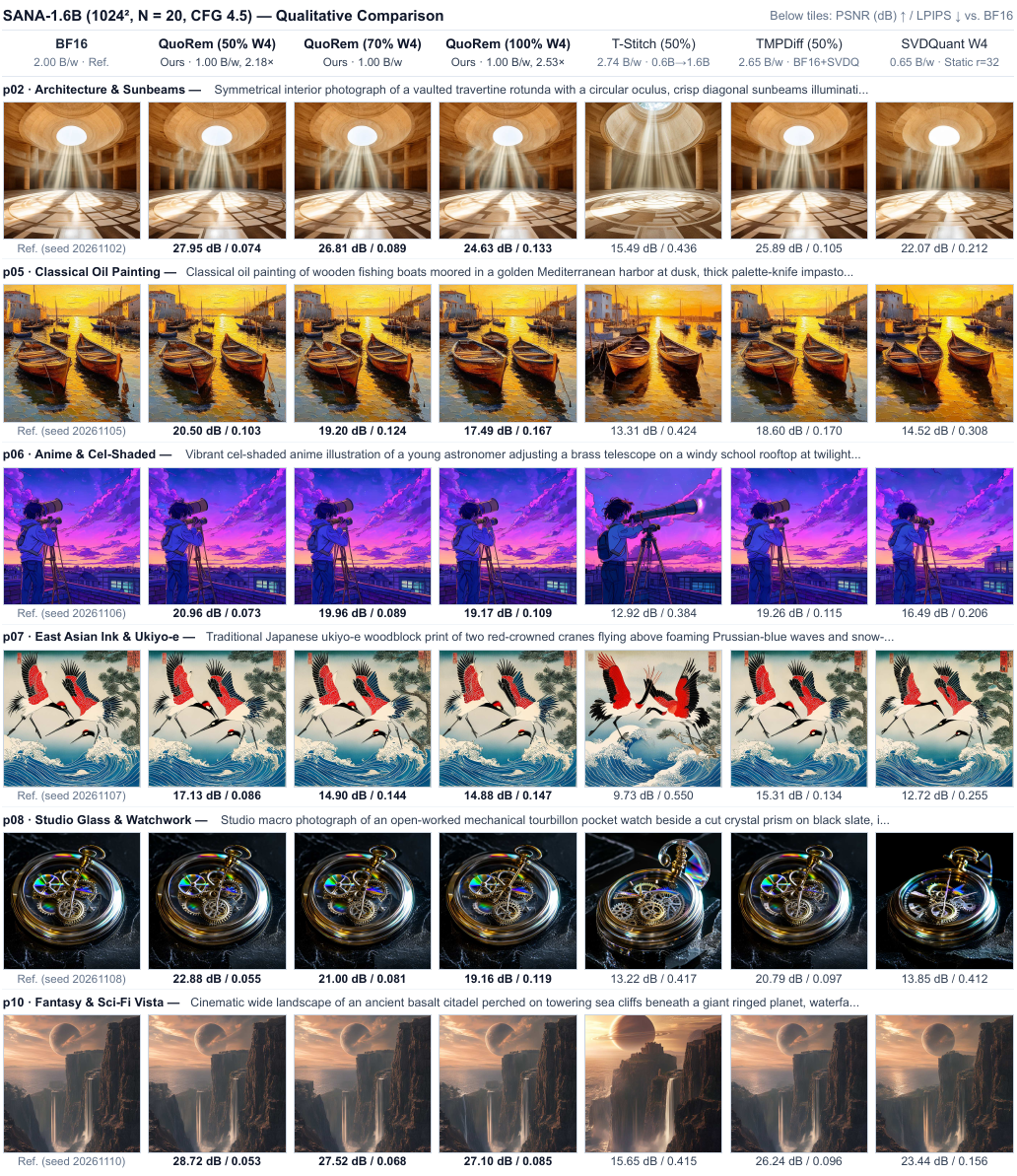}
\caption{\textbf{Full-page $1024\times 1024$ qualitative gallery on SANA-1.6B} ($N=20$, CFG $4.5$) across six visual categories. \textbf{QuoRem} ($r=0$, $1.00\text{ B/w}$) retains travertine sunbeams, palette-knife impasto, cel-shaded linework, ukiyo-e keylines, tourbillon caustics, and alpine vistas through $\tau=70\%$--$100\%$, whereas \ts\ ($\text{0.6B}\to\text{1.6B}$, $2.74\text{ B/w}$) alters scene geometry and \td\ ($2.65\text{ B/w}$) and static SVDQuant W4 lose fine contrast.}
\label{fig:supp_s2}
\end{figure*}

\begin{figure*}[!p]
\centering
\includegraphics[width=\textwidth]{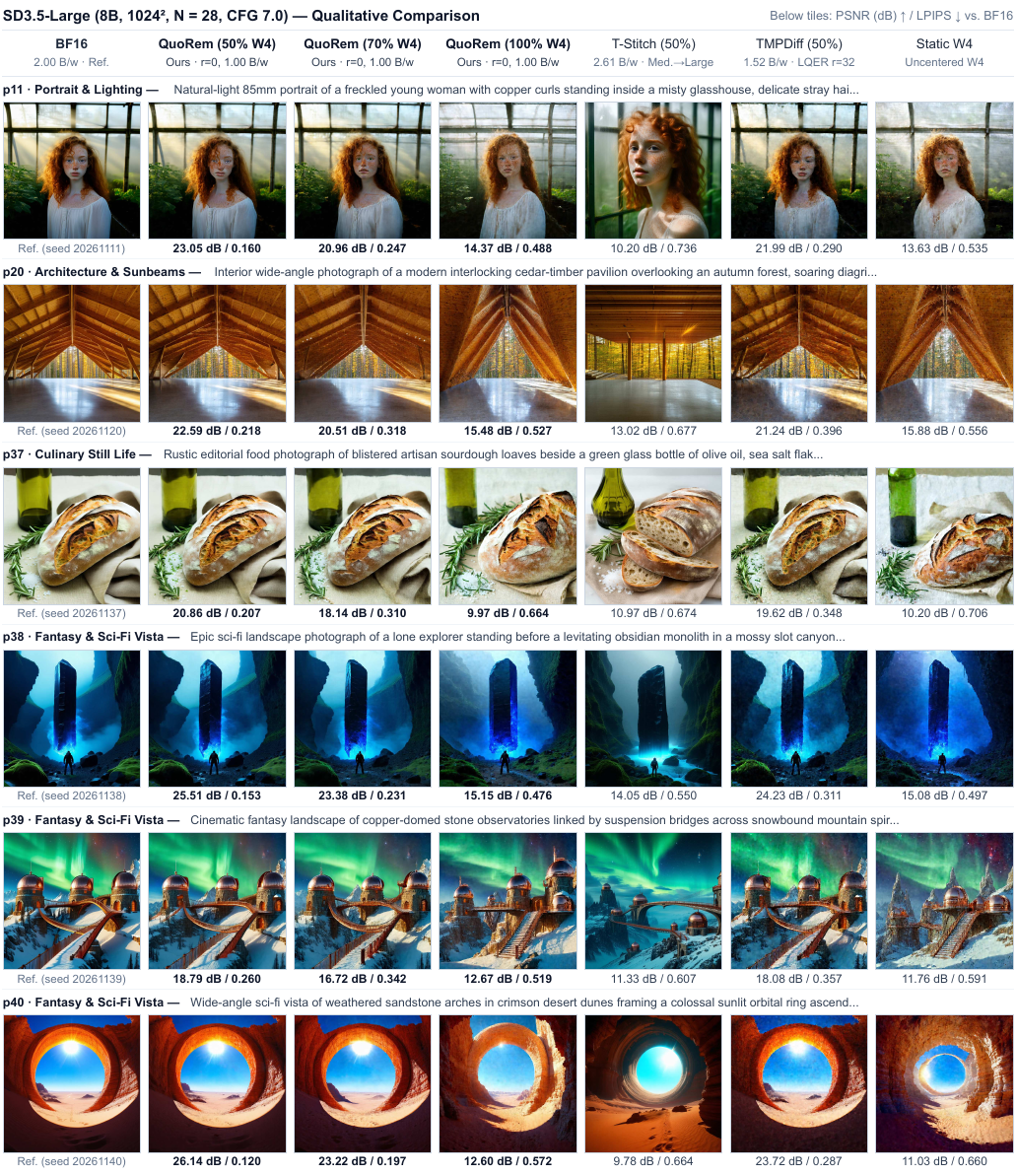}
\caption{\textbf{Full-page $1024\times 1024$ qualitative gallery on SD3.5-Large} ($8$B, $N=28$, CFG $7.0$). Across portraits, architectural diagrids, culinary close-ups, and sci-fi/fantasy vistas, \textbf{QuoRem} ($r=0$, $1.00\text{ B/w}$, Waterline schedule) avoids both the severe layout and pose shifts of cross-model \ts\ ($\text{Medium}\to\text{Large}$, $2.61\text{ B/w}$) and the grainy surface artifacts of dual-checkpoint \td\ ($1.52\text{ B/w}$, rank-$32$ LQER) and uncentered Static W4.}
\label{fig:supp_s3}
\end{figure*}

\begin{figure*}[!p]
\centering
\includegraphics[width=\textwidth]{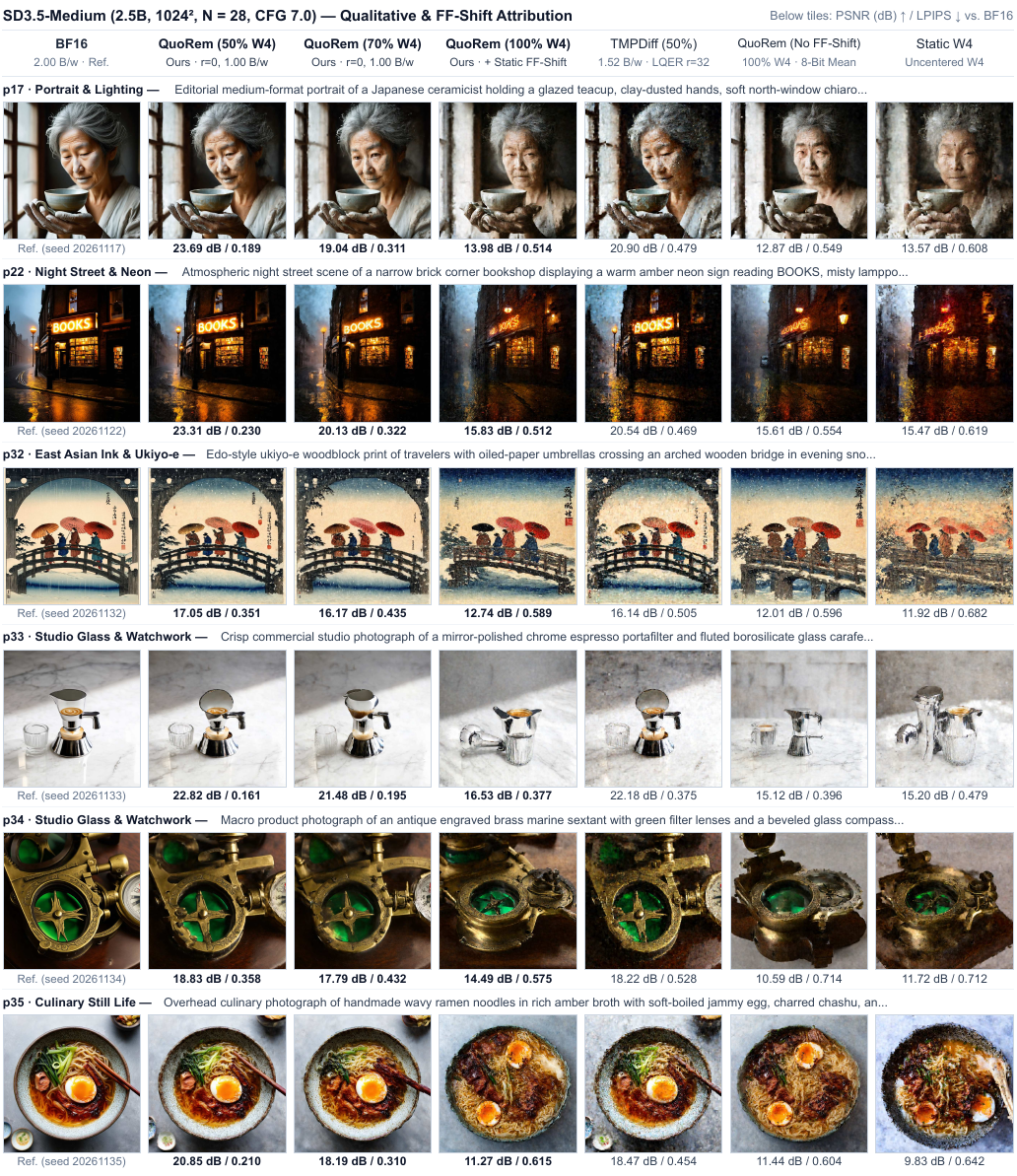}
\caption{\textbf{Full-page $1024\times 1024$ qualitative gallery and Static FF-Shift attribution on SD3.5-Medium} ($2.5$B, $N=28$, CFG $7.0$). \textbf{QuoRem} ($r=0$, $1.00\text{ B/w}$) maintains crisp skin chiaroscuro, neon typography (\texttt{BOOKS}), woodblock snow gradations, metallic specular reflections, and culinary textures at $\tau=50\%$ and $\tau=70\%$ over \td\ ($1.52\text{ B/w}$), while +\,Static FF-Shift at $\tau=100\%$ prevents the post-GELU bias drift seen in QuoRem (No FF-Shift) and uncentered Static W4.}
\label{fig:supp_s4}
\end{figure*}

\begin{figure*}[!p]
\centering
\includegraphics[width=\textwidth]{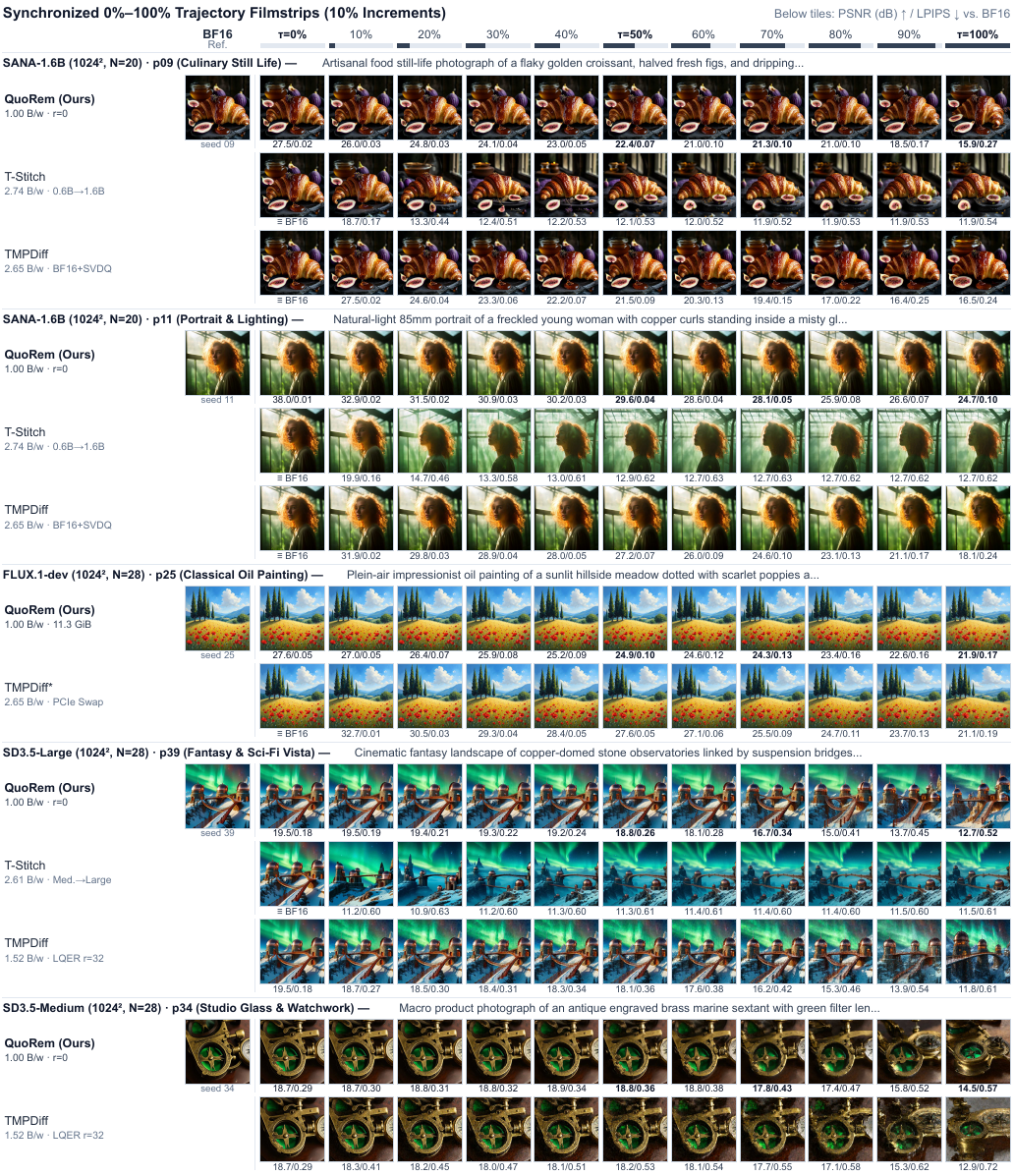}
\caption{\textbf{Synchronized $0\%$--$100\%$ trajectory filmstrips in $10\%$ W4 increments} across SANA-1.6B (\texttt{p09}, \texttt{p11}), FLUX.1-dev (\texttt{p25}; no \ts), SD3.5-Large (\texttt{p39}), and SD3.5-Medium (\texttt{p34}). Single-parent \textbf{QuoRem} transitions smoothly from its W8 parent view ($\tau=0\%$) through $\tau=70\%$ without structural jumps, whereas \ts\ shifts object pose by $\tau=10\%$--$20\%$ and \td\ degrades earlier in texture sharpness.}
\label{fig:supp_s5}
\end{figure*}

\begin{figure*}[!p]
\centering
\includegraphics[width=\textwidth]{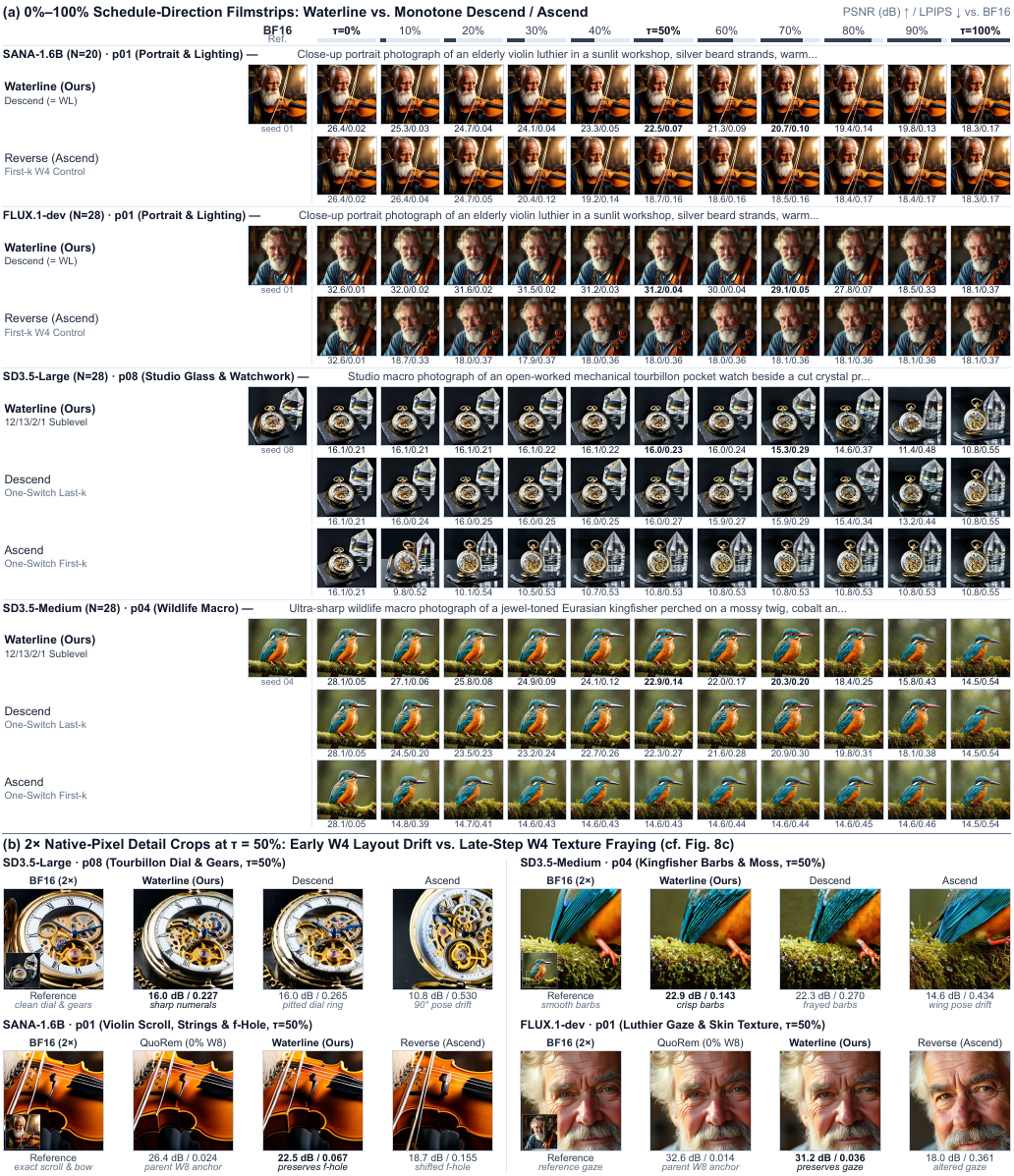}
\caption{\textbf{Schedule-direction filmstrips ($0\%$--$100\%$) and $2\times$ native-pixel detail attribution at $\tau=50\%$} across all four DiTs (extending Fig.~\ref{fig:waterline_valid}c). \textbf{(a)}~Full $11$-stop sweeps comparing \textbf{Waterline} against monotone \emph{Descend} and \emph{Ascend} (\emph{Reverse}). \textbf{(b)}~$2\times$ crops at $\tau=50\%$: early-W4 \emph{Ascend}/\emph{Reverse} induces global pose, f-hole, and gaze drift, while naive late-W4 \emph{Descend} on SD3.5-L/M pits watch dials and frays feather barbs at step $27$; \textbf{Waterline} preserves both layout and fine detail.}
\label{fig:supp_s6}
\end{figure*}

Figures~\ref{fig:supp_s1}--\ref{fig:supp_s6} provide full-page $1024\times 1024$ qualitative galleries, synchronized $0\%$--$100\%$ trajectory filmstrips, and schedule-direction case analyses across all four DiT backbones under fixed random seeds.
All visual panels report paired PSNR~(dB)~$\uparrow$ and LPIPS~$\downarrow$ against the same-seed BF16 reference.

\paragraph{Backbone galleries and component attribution (Figs.~\ref{fig:supp_s1}--\ref{fig:supp_s4}).}
Across FLUX.1-dev (Fig.~\ref{fig:supp_s1}), SANA-1.6B (Fig.~\ref{fig:supp_s2}), SD3.5-Large (Fig.~\ref{fig:supp_s3}), and SD3.5-Medium (Fig.~\ref{fig:supp_s4}), single-parent \textbf{QuoRem} ($r=0$, $1.00\text{ B/w}$) preserves global composition, lighting, specular caustics, and fine typography at $\tau=50\%$ and $\tau=70\%$ W4 step shares without a second checkpoint.
On FLUX.1-dev (no smaller sibling architecture exists for \ts; Tab.~\ref{tab:systems}b), QuoRem runs in $11.29\text{ GiB}$ ($1.64\times$ BF16 speed at $\tau=50\%$), avoiding the $29.6\text{ GiB}$ co-residency overflow and PCIe swaps of dual-checkpoint \td$^\ast$ while avoiding the structural washout of static SVDQuant W4 and uncentered Static W4.
On SANA-1.6B and SD3.5-Large, sibling stitching (\ts) alters camera framing, object pose, and architectural geometry because early prefix steps execute a distinct smaller backbone (Sec.~\ref{sec:deletion}, Fig.~\ref{fig:stitch_align}), whereas \td\ introduces mid-frequency grain from uncentered 4-bit activations.
On SD3.5-Medium (Fig.~\ref{fig:supp_s4}), comparing QuoRem ($100\%$ W4) against QuoRem without Static FF-Shift isolates Proposition~\ref{prop:mean}: folding calibrated post-GELU shifts into layer biases removes one-sided activation skew and restores sign legibility, metallic specular reflections, and skin chiaroscuro at zero online reduction cost.

\paragraph{Continuous $0\%$--$100\%$ trajectory filmstrips and schedule attribution (Figs.~\ref{fig:supp_s5}--\ref{fig:supp_s6}).}
Sweeping the W4 step share $\tau$ from $0\%$ to $100\%$ in $10\%$ increments (Fig.~\ref{fig:supp_s5}) shows smooth, monotone fidelity transitions under QuoRem up to $\tau\approx 70\%$, whereas \ts\ jumps to an altered scene layout as soon as $10\%$--$20\%$ of early steps run the draft model.
Finally, Fig.~\ref{fig:supp_s6} verifies the Euler--Wiener survival analysis of Lemma~\ref{lem:ew_gain} and Theorem~\ref{thm:direction} (complementing Fig.~\ref{fig:waterline_valid}c): early-W4 schedules (\emph{Ascend} / \emph{Reverse}) inject uncontracted layout perturbations that rotate watch dials, shift violin f-holes, and alter gaze, while naive late-W4 \emph{Descend} on SD3.5-Large/Medium leaves large tail step $27$ in W4, pitting enamel rings and fraying kingfisher feather barbs.
\textbf{Waterline} ($12/13/2/1$ on SD3.5-M/L; coinciding with Descend on SANA/FLUX where online 8-Bit Mean suppresses tail drift) protects both high-gain boundaries and reclaims tiny final step $28$ in W4, preserving layout and crisp high-frequency texture simultaneously.

\end{document}